\UseRawInputEncoding

\documentclass[journal]{IEEEtran}  

\usepackage{multirow}
\usepackage[table,xcdraw]{xcolor}

\usepackage{amsmath,amsfonts}
\usepackage{amsmath, amssymb}
\usepackage{bm}
\usepackage{algorithmic}
\usepackage{array}
\usepackage[caption=false,font=footnotesize,labelfont=rm,textfont=rm]{subfig}
\usepackage{textcomp}
\usepackage{stfloats}
\usepackage{url}
\usepackage{verbatim}
\usepackage{graphicx}

\usepackage{color}
\usepackage{amssymb}
\usepackage{booktabs}
\usepackage[table]{xcolor}
\usepackage{graphicx}
\usepackage{multirow}
\usepackage{color} 
\usepackage{cite}
\usepackage{mathrsfs}

\newcommand{\figref}[1]{Fig. \ref{#1}}

\newcommand{\ra}[1]{\renewcommand{\arraystretch}{#1}}

\newcommand{\tabref}[1]{Table \ref{#1}}

\newcommand{\steer}{\mathbf{a}}
\newcommand{\mean}{\mathbb{E}}

\usepackage{algorithm}
\usepackage{algorithmic}

\newcommand{\alref}[1]{Algorithm \ref{#1}}

\makeatletter
\g@addto@macro\normalsize{%
	\setlength\abovedisplayskip{3pt} 
	\setlength\belowdisplayskip{3pt}  
	\setlength\abovedisplayshortskip{1.5pt} 
	\setlength\belowdisplayshortskip{1.5pt} 
	\setlength\textfloatsep{3.1pt}  
	\setlength{\abovecaptionskip}{0pt}  
	\setlength{\belowcaptionskip}{0pt} 
}
\makeatother

\newcommand{\subparagraph}{}
\usepackage{titlesec}
\titlespacing{\section}{0pt}{1.5ex plus .3ex minus .1ex}{.5ex plus .1ex}
\titlespacing{\subsection}{0pt}{1.5ex plus .3ex minus .1ex}{.5ex plus .1ex}  
\allowdisplaybreaks[4]

\def\BibTeX{{\rm B\kern-.05em{\sc i\kern-.025em b}\kern-.08em
		T\kern-.1667em\lower.7ex\hbox{E}\kern-.125emX}}
\usepackage{balance}
\usepackage{hyperref}
\hypersetup{hidelinks} 
\hypersetup{
	    colorlinks=true,
		linkcolor=blue
	}

\begin{document}
\title{Statistical Channel Fingerprint Construction for Massive MIMO: A Unified Tensor Learning Framework}

\author{
	Zhenzhou~Jin,~\IEEEmembership{Graduate Student Member,~IEEE,}
	Li~You,~\IEEEmembership{Senior Member,~IEEE,}\\
	Xiang-Gen~Xia,~\IEEEmembership{Fellow,~IEEE,}
	and~Xiqi~Gao,~\IEEEmembership{Fellow,~IEEE}
	\thanks{Part of this work has been accepted for presentation at the IEEE VTC2026-Spring \cite{jinvtc}.}	
	\thanks{
		Zhenzhou Jin, Li You, and Xiqi Gao are with the National Mobile Communications Research Laboratory, Southeast University, Nanjing 210096, China, and also with the Purple Mountain Laboratories, Nanjing 211100, China (e-mail: zzjin@seu.edu.cn; lyou@seu.edu.cn; xqgao@seu.edu.cn).
		
%
%
		Xiang-Gen Xia is with the Department of Electrical and Computer Engineering, University of Delaware, Newark, DE 19716, USA (e-mail: xxia@ee.udel.edu).
	}



			
	}

\maketitle

\begin{abstract}
Channel fingerprint (CF) is considered a key enabler for facilitating the acquisition of channel state information (CSI) in massive multiple-input multiple-output (MIMO) communication systems. In this work, we investigate a novel type of CF that stores statistical CSI (sCSI) at each potential location, referred to as statistical CF (sCF). Specifically, we reveal the relationship between sCSI, namely the channel spatial covariance matrix (CSCM), and the channel power angular spectrum (CPAS). Building on this foundation, we construct a unified tensor representation of the sCF and further reduce its dimension by exploiting the eigenvalue decomposition of the CSCM and its correlation with the PAS. Considering the practical constraints imposed by measurement cost, privacy, and security, we focus on three representative scenarios and uniformly formulate them as tensor restoration tasks. To this end, we propose a unified tensor-based learning architecture, termed LPWTNet. The architecture incorporates a closed-form Laplacian pyramid (LP) decomposition and reconstruction framework that replaces the traditional encoder-decoder structure, enabling efficient inference while capturing multi-scale frequency subband characteristics of the sCF. Additionally, a shared mask learning strategy is introduced to adaptively refine high-frequency sCF components through level-wise adjustments. To achieve a larger receptive field without over-parameterization, we further propose a small-kernel convolution mechanism based on the wavelet transform (WT), which decouples convolution across different frequency components of the sCF and enhances feature extraction efficiency. Extensive experiments show that the proposed approach delivers competitive reconstruction accuracy and computational efficiency across various sCF construction scenarios when compared with state-of-the-art baselines.

\end{abstract}


\section{Introduction}\label{sec:net_intro}

\IEEEPARstart{T}{he} convergence of artificial intelligence (AI) and wireless communications has emerged as a highly transformative paradigm for next-generation mobile communication networks \cite{10054381,11424370}. The forthcoming sixth-generation (6G) communication networks are expected to achieve ultra-high performance metrics while supporting emerging scenarios such as the intelligent internet of everything and ubiquitous connectivity. Consequently, 6G will demand significantly enhanced end-to-end information processing capabilities. In this context, the seamless integration of AI into wireless communication systems is increasingly recognized as a pivotal enabler, allowing the overcoming of traditional performance limitations and fundamentally reshaping the wireless network ecosystem \cite{8808168}.

With the rapidly growing antenna array dimensions in massive multiple-input multiple-output (MIMO) communication systems, along with the explosive increase in the number of connected devices and the adoption of higher frequency bands, 6G will inevitably face the challenge of handling ultra-high-dimensional MIMO channels \cite{10403776,jin2024i2i,10920478}. Channel state information (CSI) at the base station (BS) plays a critical role in massive MIMO transmission. In practical systems, it is typically obtained through periodically inserted pilot signals. However, as the dimension of MIMO channels escalates, traditional pilot-based CSI acquisition and feedback mechanisms impose prohibitively high pilot signaling overhead \cite{reuse}. Channel fingerprint (CF), also referred to as the channel knowledge map (CKM), has recently emerged as a promising enabler for facilitating CSI acquisition by providing location-specific channel knowledge for any BS-to-everything (B2X) pair \cite{zeng2024tutorial,jin2024i2i}. More specifically, a CF can be regarded as a knowledge repository that stores channel-related information, such as angles of arrival and departure, channel power, and even channel impulse responses, for user terminals (UTs) at each potential location within the area of interest \cite{11220249}.

\subsection{Prior Work}
By providing accurate and essential channel-related knowledge, various types of CFs have been widely applied in integrated air-ground-sea networks to support global and seamless coverage. Existing applications include beamforming \cite{wu2023environment, zeng2024tutorial}, assisting the establishment of links between non-cooperative nodes \cite{zeng2024tutorial}, communication and path planning for low-altitude unmanned aerial vehicles (UAVs) \cite{11224604, 9269485}, wireless physical environment sensing \cite{zhang2020constructing, 9838964}, UT localization \cite{yang2013rssi}, resource allocation in massive MIMO systems \cite{10556774, sun2017bdma, li2021downlink}.

However, the effectiveness of the aforementioned CF-enabled wireless transmission can be fundamentally ensured only when CFs are accurately constructed. To this end, prior studies have made valuable contributions by exploring different scenarios and various types of CFs. Existing works can generally be categorized into two groups based on their construction methodologies: model-driven and data-driven. In the model-driven category, the authors of \cite{10530520} propose an analytical channel gain model to characterize the spatial variation of the wireless environment, using partially measured channel gain information to estimate the channel gain fingerprint. In \cite{8662745}, a piecewise homogeneous spatial loss field model is introduced, in which the environment is divided into regions with uniform attenuation, and the path loss at unmeasured locations is inferred using prior assumptions about the environment. Furthermore, the authors of \cite{10634565} focus on channel power fingerprints and develop a method that integrates interpolation with block-term tensor decomposition to utilize sparsely measured channel power data.

On the data-driven side, the authors of \cite{zhang2024radiomap} focus on predicting received signal strength and introduce a two-step template-perturbation method for inpainting measurements in inaccessible regions. Additionally, AI-based methods for CF construction have recently gained increasing attention. In \cite{zeng2024tutorial}, fully connected networks are adopted to estimate channel gain fingerprints from two-dimensional coordinate inputs. In \cite{jin2024i2i}, the CF construction task is reformulated as an image-to-image (I2I) inpainting problem, and an efficient frequency-aware network is developed to enhance reconstruction performance. In \cite{9354041,9771088}, U-Net is adopted to extract geometry-related features in urban or indoor environments, while sparsely measured channel path gains are utilized to enable the construction of corresponding CFs. Moreover, with recent advances in generative AI, diffusion models and their conditional variants have been introduced to construct different types of CFs from sparse measurement data \cite{11220249,11190081}.

\subsection{Motivations and Contributions}
Accurate construction of CFs is essential for enabling optimized wireless transmission design. However, most existing studies focus on constructing CFs that store only a single type of channel parameter and typically represent instantaneous CSI (iCSI). These works also tend to consider only a single CF construction scenario, such as reconstructing CFs from uniformly and sparsely measured data. Such approaches may not effectively align with real-world communication scenarios for two key reasons: \textit{i)} In complex and dynamic environments, obtaining reliable iCSI becomes increasingly challenging. Additionally, due to computational complexity and limitations in data representation, CFs that store only a single type of channel-related parameter exist in a simple scalar form at each UT position, which may limit their applicability across diverse scenarios. \textit{ii)} Due to the high cost of measurements and sensing, as well as privacy and security concerns, practical CF construction scenarios are highly diverse and often involve typical tasks such as non-uniform sparse recovery, missing-region estimation, and uniform sparse recovery. Moreover, long-term channel statistics, which vary more slowly over time, are inherently more robust within a given time-slot window and capture a broader range of channel-related statistical properties compared to iCSI \cite{adhikary2013joint,5165196}. Numerous studies have also demonstrated that statistical CSI (sCSI) can effectively support critical transmission applications, such as pilot reuse, precoding design, and power allocation, in massive MIMO systems \cite{reuse,sun2017bdma,li2021downlink,adhikary2013joint,5165196}.

Motivated by the above discussion, this paper introduces a CF that stores sCSI, referred to as the statistical CF (sCF). The sCF is formulated as a multi-dimensional tensor, in which each spatial location contains a vectorized channel knowledge representation that conveys richer and more comprehensive channel-related information compared with the scalar values adopted in existing CF construction methods. Moreover, we consider several typical sCF reconstruction scenarios and propose a unified tensor-based learning architecture, termed LPWTNet, to handle these diverse conditions. The main contributions of this paper are summarized as follows:
\begin{itemize}
	\item 
	\textbf{Unified tensor-based sCF model and multi-scenario sCF construction problem formulation}: Within the proposed massive MIMO statistical channel model, we reveal the relationship between sCSI, specifically the channel spatial covariance matrix (CSCM), and the channel power angular spectrum (CPAS). Building on this, we model the target sCF using a unified tensor representation by vectorizing the CSCM at each possible UT location. Furthermore, by leveraging the eigenvalue decomposition of the CSCM and its correlation with the PAS, we efficiently reduce the dimension of the sCF in tensor form. To accommodate diverse real-world scenarios for sCF construction, we consider three typical cases within the proposed sCF model, and the corresponding problems are uniformly formulated as tensor restoration tasks.

	\item \textbf{Efficient multi-scale feature decomposition and reconstruction}: We introduce a Laplacian pyramid (LP) framework with closed-form frequency-band decomposition and reversible reconstruction to replace the traditional encoder-decoder architecture. This framework not only mitigates the substantial computational overhead and irreversible information loss of conventional designs but also enables a multi-scale frequency subband representation of the sCF. To ensure consistent restoration of detailed information in the high-frequency sCF components, we incorporate the global structural priors contained in the low-frequency components and propose a shared mask learning strategy that performs level-wise adaptive adjustments to refine the high-frequency components. 		
	\item 
	\textbf{Enhanced receptive fields through wavelet-domain small-kernel convolutions and portable module design}: Instead of using large kernels or high-complexity attention to enlarge the receptive field, we propose an efficient feature extraction mechanism based on wavelet-domain small-kernel convolution, termed wavelet transform convolution (WTConv). By combining the WT with depthwise separable convolution, this mechanism decouples convolution operations across different frequency components and allows small kernels to effectively cover a larger spatial region of the original input. Furthermore, to enable the model to adaptively modulate the importance of frequency components at different decomposition levels, we introduce a wavelet scale layer after each stage of wavelet decomposition. Building on these components, we develop the Res-DSWT block, which can be seamlessly integrated into diverse neural network architectures.
	\item 
	\textbf{Experimental validation of key modules and the proposed approach}: The ablation experiments demonstrate that the proposed Res-DSWT block yields substantial performance gains in sCF construction tasks compared with conventional convolutional modules and self-attention mechanisms. In addition, the proposed LPWTNet is evaluated against several baseline models across three representative sCF reconstruction scenarios, showing competitive performance in terms of both computational complexity and reconstruction accuracy.
		
\end{itemize}

\subsection{Organization and Notations}
The rest of this paper is structured as follows. Section \ref{sec:sys_mod} introduces the system and sCF model, along with the unified formulation of the sCF construction problem across multiple scenarios. The overall design of LPWTNet, including its specific modules and network architecture, is presented in Section \ref{sec:LPWTNet-Based sCF Constru}. Experimental results are provided in Section \ref{sec:Numerical Experiment}, followed by the conclusions in Section \ref{sec:conclusion}. The main notations used in this paper are listed in \tabref{notation} for reference.
\newcolumntype{L}{>{\hspace*{-\tabcolsep}}l}
\newcolumntype{R}{c<{\hspace*{-\tabcolsep}}}
\definecolor{lightblue}{rgb}{0.93,0.95,1.0}
\begin{table}[htbp]
	\captionsetup{font=footnotesize}
	\caption{Notation List}\label{notation}
	\centering
	\setlength{\tabcolsep}{12.6mm}
	\ra{1.6}
	\scriptsize
	\scalebox{0.8}{\begin{tabular}{LR}
			\toprule
			Notation &   Definition\\
			\midrule
			\rowcolor{lightblue}
			$\bar\jmath=\sqrt { - 1} $ & Imaginary unit   \\
			$\otimes$ & Kronecker product    \\
			\rowcolor{lightblue}
			$\triangleq$ & Definition\\
			$\mean\{\}$ & Expectation\\
			\rowcolor{lightblue}
			$\delta(\cdot)$& Dirac delta function\\
			${\rm diag} \left\lbrace\steer  \right\rbrace$ & Diagonalization operator\\
			\rowcolor{lightblue}
			$\mathbf{A}$ & Matrix\\
			$\steer$ & Vector\\
			\rowcolor{lightblue}
			$\boldsymbol{\mathcal{F}}$ & 3D tensor\\
			$(\cdot)^H$ & Conjugate transpose\\
			\rowcolor{lightblue}
			$\mathbb{R}^{N\times M}$ & Real-valued matrix space of size $N\times M$\\
			$\mathbb{C}^{N\times M}$ & Complex-valued matrix space of size $N\times M$\\
			\rowcolor{lightblue}
			$\circ$ & Hadamard product\\
			$\mathcal{M}: \boldsymbol{\mathcal{X}}\to \boldsymbol{\mathcal{Y}}$ & Mapping from $\boldsymbol{\mathcal{X}}$ to $\boldsymbol{\mathcal{Y}}$\\
			\rowcolor{lightblue}
			$||\cdot||$  & Euclidean norm\\
			$\mathbf{A}_{i,j}$ & the $(i,j)$-th element of matrix $\mathbf{A}$\\
			\rowcolor{lightblue}
			$\boldsymbol{\mathcal{F}}_{i,j,:}$ & The vector along the third dimension at row $i$ and column $j$\\
			$\boldsymbol{\mathcal{F}}_{:,:,c}$ &  The 2D slice of $\boldsymbol{\mathcal{F}}$ at the $c$-th channel (i.e., the third dimension)\\
			\rowcolor{lightblue} $\boldsymbol{\mathcal{F}}_{i,j,c}$ &
			The element of the tensor 
			$\boldsymbol{\mathcal{F}}$ at row $i$, column $j$, and channel $c$\\
			$(\cdot)^T$ & Transpose operator\\
			\rowcolor{lightblue}
			$\mathbb{R}^{N\times M \times C}$  & Real-valued tensor space of size $N\times M\times C$\\
			$\mathbb{C}^{N\times M \times C}$ & Complex-valued tensor space of size $N\times M\times C$ \\
			\rowcolor{lightblue}
			$(\cdot)^*$ & Conjugate operation\\
			$\circledast$ & 2D convolution operation \\
			\rowcolor{lightblue}
			a mod b & Modulo operation\\
			$\lceil \cdot \rceil$ &   Ceiling function\\
			\bottomrule
		\end{tabular}
	}
\end{table}

\section{System Model And Problem Formulation}\label{sec:sys_mod}
In this section, we first introduce the wireless communication scenario considered in this work and present the massive MIMO statistical channel model for each potential UT location. We then analyze the sCSI by characterizing the relationship between the CSCM and the CPAS. By exploiting this intrinsic correlation, a low-dimensional sCF model is constructed for the geometric region of interest. Finally, the sCF construction tasks arising in three practical scenarios are formulated from a unified tensor-degeneration perspective.

\subsection{Massive MIMO Statistical Channel Model}\label{subsection:channel}
Consider a massive MIMO communication system operating within a square region $A\subset {\mathbb{R}^2}$, where the locations of potential UTs are denoted by the set $\left\{ {{{{\mathbf{x}}_k}}} \right\}_{k = 1}^K = {\mathcal{K}}$. Specifically, the BS is equipped with a uniform planar array (UPA) consisting of $N=N_{y}\times N_{z}$ antenna elements and serves $K$ single-antenna UTs over frequency-flat fading channels on a narrowband subcarrier. In this case, the azimuth and elevation angles at the BS side are denoted by $\varphi $ and $\theta $, respectively. Without loss of generality, based on the ray-tracing approach \cite{reuse,barriac2004characterizing}, the channel between the BS and the UT located at ${{{{\mathbf{x}}_k}}}$ can be modeled as
\begin{align}\label{eq:1}
	\mathbf{h}_{{{{{\mathbf{x}}_k}}}} = \int\limits_{\theta_{\rm min}}^{\theta_{\rm max}} \int\limits_{\varphi_{\rm min}}^{\varphi_{\rm max}}  g_{{{{{\mathbf{x}}_k}}}} (\theta, \varphi)   \steer(\theta, \varphi ) d\varphi d\theta \in \mathbb{C}^{N\times 1},
\end{align}
where $ \steer(\theta, \varphi ) \in \mathbb{C}^{ N \times 1} $ and $ g_{{{{{\mathbf{x}}_k}}}}(\theta, \varphi ) $ denote the steering vector and the complex-valued channel gain, respectively. With the following definitions
\begin{align}\label{eq:}
	{{\mathbf{a}}_y}\left( {\theta ,\varphi } \right) &= {\left[ {1,{e^{ - \bar \jmath 2\pi \upsilon \cos \theta \sin \varphi }}, \ldots ,{e^{ - \bar \jmath 2\pi (N_y - 1)\upsilon \cos \theta \sin \varphi }}} \right]^T}, \\
	{{\mathbf{a}}_z}\left( \theta  \right) &= {\left[ {1,{e^{ - \bar \jmath 2\pi \upsilon \sin \theta  }}, \ldots ,{e^{ - \bar \jmath 2\pi ({N_z} - 1)\upsilon \sin \theta }}} \right]^T},
\end{align}
the steering vector $\steer(\theta, \varphi )$ in \eqref{eq:1} can be obtained via \cite{reuse}
\begin{align}\label{eq:}
	\steer(\theta, \varphi ) = \steer_{z}(\theta ) \otimes \steer_{y}(\theta, \varphi),
\end{align}
where $\upsilon  \triangleq d/\lambda$, $d\triangleq\lambda/2$ is the inter-antenna spacing, and $\lambda$ denotes the wavelength. Based on the introduced channel model in \eqref{eq:1}, the azimuth and elevation angles of the channel are discretized into $ \{ \varphi(n) \}_{n=0}^{N-1} $ and $ \{ \theta(n) \}_{n=0}^{N-1} $, respectively, where $ \varphi(n) \in \left[ \varphi_{\rm min}, \varphi_{\rm max} \right] $ and $\theta(n) \in \left[ \theta_{\rm min}, \theta_{\rm max} \right]$. Let
\begin{align}
	\mathbf{A} &= \left[ \steer\left( \theta(0), \varphi(0) \right) , \dots, \steer\left( \theta(N-1), \varphi(N-1) \right)   \right] , \\
	\mathbf{g}_{ {{{{\mathbf{x}}_k}}}} &= \left[  g_{ {{{{\mathbf{x}}_k}}}}(\theta(0), \varphi(0)), \dots, g_{ {{{{\mathbf{x}}_k}}}}(\theta(N-1), \varphi(N-1)) \right]^{T},
\end{align}
then the channel \eqref{eq:1} can be approximated as
\begin{align}
	\mathbf{h}_{{{{{\mathbf{x}}_k}}}} = \mathbf{A} \mathbf{g}_{{{{{\mathbf{x}}_k}}}}.
\end{align}
Assuming the channel phase follows a uniform distribution and channels with distinct incident angles are mutually uncorrelated \cite{pesersen,reuse}, we have
\begin{align}\label{eq:s}
	\mean\left\lbrace   g_{{{{{\mathbf{x}}_k}}}}(\theta(n), \varphi(n)) g_{{{{{\mathbf{x}}_k}}}}^{*}(\theta(n'), \varphi(n'))  \right\rbrace  \!\!=\! {\xi _{{{\mathbf{x}}_k}}}S_{{{{{\mathbf{x}}_k}}}}(n) \delta( n\!-\!n'),
\end{align} 
where ${\xi _{{{\mathbf{x}}_k}}}$ represents the large-scale fading coefficient, and the sequence $\mathcal{S}_{{{{{\mathbf{x}}_k}}}} \triangleq \{{\xi _{{{\mathbf{x}}_k}}}S_{{{{{\mathbf{x}}_k}}}}(n)\}_{n=1}^{N}$ characterizes the power distribution of the channel in the angular domain \cite{pesersen}.

In light of the above analysis, the CSCM corresponding to the UT located at ${{{{\mathbf{x}}_k}}}$ is given by 
\begin{subequations}\label{eq:r}
\begin{align}
	\bm{\Omega}_{{{{\mathbf{x}}_k}}} &= \mean\left\lbrace \mathbf{h}_{{{{{\mathbf{x}}_k}}}} \mathbf{h}_{{{{{\mathbf{x}}_k}}}}^{H} \right\rbrace \\
	&= \mathbf{A}  \mean\left\lbrace \mathbf{g}_{{{{{\mathbf{x}}_k}}}} \mathbf{g}_{{{{{\mathbf{x}}_k}}}}^{H} \right\rbrace \mathbf{A}^{H}  \\
	&\overset{(\mathrm{a})}{=} \mathbf{A} {\rm diag} \left\lbrace  \mathcal{S}_{{{{\mathbf{x}}_k}}}\right\rbrace  \mathbf{A}^{H}  \\
	&= \mathbf{A} \widetilde{\bm{\Omega}}_ {{{\mathbf{x}}_k}} \mathbf{A}^{H} ,	
\end{align}
\end{subequations}
where $(\mathrm{a})$ follows from \eqref{eq:s}, and $ \widetilde{\bm{\Omega}}_ {{{\mathbf{x}}_k}} \triangleq  {\rm diag} \left\lbrace  \mathcal{S}_{{{{\mathbf{x}}_k}}}\right\rbrace \in \mathbb{R}^{N\times N}$ represents the CPAS. \eqref{eq:r} reveals the relationship between the CSCM and the CPAS. Then, we consider the properties of the matrix $\mathbf{A}$ for further analysis. It has been shown in \cite{adhikary2013joint,6415397,reuse,huang2018joint,5165196} that, as $N$ grows large, there exists an approximately one-to-one mapping between the elevation angle $\theta(n)$ and the index $n$, i.e., $\sin \theta(n) = (2n/N) - 1$. A similar relationship holds for the azimuth angle $\varphi(n)$. Therefore, when $d = \lambda / 2$, the elements of $\mathbf{A}$ can be further expressed as
\begin{align}
	\mathbf{A}_{m,n} 
	&= \displaystyle{
		e^{ - \bar{\jmath} 2 \pi \frac{ \left( m'-1\right)  \left( n-1- \frac{N}{2}\right)}  {N} }     
		e^{ - \bar{\jmath} 2 \pi \frac{ \left( m''-1\right)  \left( n-1-\frac{N}{2}\right) } { N} }
	} \nonumber \\
	&= \displaystyle{
		e^{- \bar{\jmath} 2 \pi \frac{ \left[  \left( m'-1\right)  \left( n-1-\frac{N}{2}\right)  \right]  + \left[  \left( m''-1\right)  \left( n-1-\frac{N}{2}\right)  \right]  } { N} } },
	\label{eq:dft}
\end{align}
where $m' = \lceil \frac{m}{N_{y}} \rceil$, $m'' = m \bmod N_{y}$. The result in \eqref{eq:dft} is critical because it indicates that the matrix $\mathbf{A}\in \mathbb{C}^{N\times N}$ asymptotically approaches the unitary two-dimensional discrete Fourier transform (2D-DFT) matrix as $N$ grows large. Similar channel spatial covariance matrix decomposition for the uniform linear arrays (ULA) case was derived in \cite{reuse,adhikary2013joint,huang2018joint,5165196}.

In this case, \eqref{eq:r} can be viewed as the eigenvalue decomposition (EVD) of $\bm{\Omega}_{{{{\mathbf{x}}_k}}}$, where the columns of $\mathbf{A}$ represent the eigenvectors, and the set of diagonal elements in $\widetilde{\bm{\Omega}}_ {{{\mathbf{x}}_k}}$, i.e., $\mathcal{S}_{{{{\mathbf{x}}_k}}}$, corresponds to the set of eigenvalues. Therefore, acquiring the sCSI of the UT located at ${{{{\mathbf{x}}_k}}}$ can be further transformed into estimating the corresponding CPAS, i.e., the eigenvalues of the CSCM, thereby significantly reducing the number of parameters to be estimated. The relationship between sCSI and the CPAS, determined by the eigenvalues of the CSCM, has been well established in the context of massive MIMO transmission systems \cite{5165196,adhikary2013joint,sun2015beam}.
\begin{figure}[!t]
	\centering
	\includegraphics[width=0.49\textwidth]{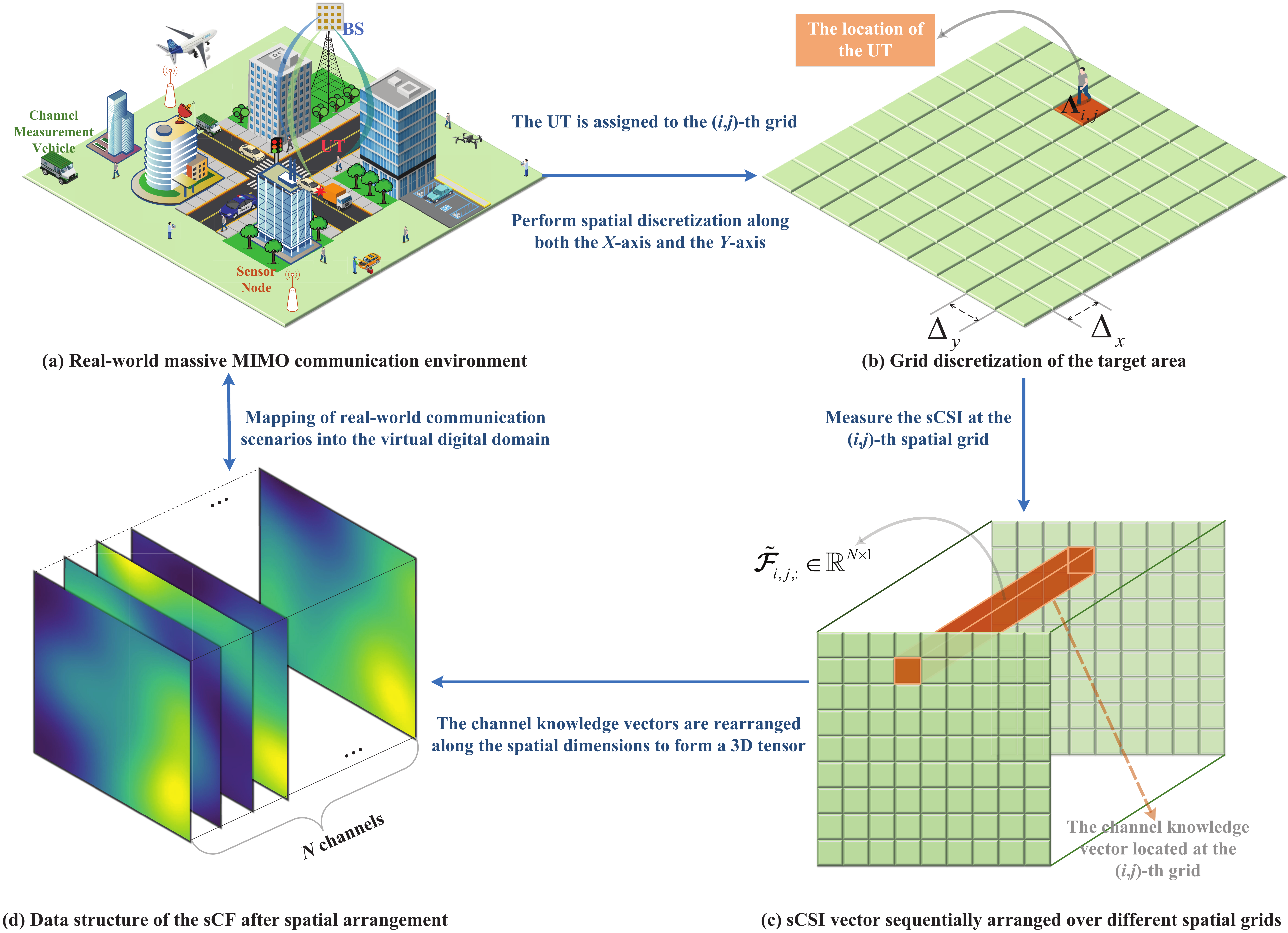}
	\captionsetup{font=footnotesize}
	\caption{Schematic diagram of the sCF modeling process}
	\label{fig:1}
\end{figure}

\subsection{Statistical Channel Fingerprints Model}\label{se:cf}
Building upon the above channel model, we proceed to model the sCF within the communication area of interest, as shown in \figref{fig:1}. Considering that the target communication area is a 2D square region $A\subset {\mathbb{R}^2}$ as specified earlier, we perform spatial discretization along both the $X$-axis and the $Y$-axis. Specifically, let the size of the target region be $\mathcal{W} \times \mathcal{W}$, and define the minimum spatial discretization units as $\Delta_x=\Delta_y={\mathcal{W} \mathord{\left/
		{\vphantom {\mathcal{W} \sigma}} \right.
		\kern-\nulldelimiterspace} \sigma}$, where $\sigma$ represents the spatial resolution. Under this setting, the $(i,j)$-th spatial grid region can be represented as
\begin{align}\label{eq:}
	{{\mathbf{\Lambda }} _{i,j}}\triangleq {[i{\Delta _x},j{\Delta {}_y]^T}},
\end{align}
where $i = 1,2,...,	{\mathcal{W} \mathord{\left/
		{\vphantom {\mathcal{W} \Delta_x}} \right.
		\kern-\nulldelimiterspace} \Delta_x}$ and $j = 1,2,...,	{\mathcal{W} \mathord{\left/
		{\vphantom {\mathcal{W} \Delta_y}} \right. \kern-\nulldelimiterspace} \Delta_y}$. Each UT located at $\left\{ {{{{\mathbf{x}}_k}}} \right\}_{k = 1}^K$ is assigned to the $(i,j)$-th spatial grid region ${{\mathbf{\Lambda }} _{i,j}}$ if and only if $\left\| {{{\mathbf{\Gamma }}_{i,j}} - {{\mathbf{x}}_k}} \right\| \leq \left\| {{{\mathbf{\Gamma }}_{i',j'}} - {{\mathbf{x}}_k}} \right\|,\forall (i',j') \ne (i,j)$, where ${{\mathbf{\Gamma }}_{i,j}}$ denotes the center of the $(i,j)$-th grid region. In the case of equal distances to grid centers ${{\mathbf{\Gamma }}_{i,j}}$ and ${{\mathbf{\Gamma }}_{i',j'}}$, the UT located at ${{\mathbf{x}}_k}$ is arbitrarily assigned to either ${{\mathbf{\Lambda }} _{i,j}}$ or ${{\mathbf{\Lambda }} _{i',j'}}$. Without loss of generality, to acquire the ground-truth sCSI at ${{\mathbf{\Lambda }} _{i,j}}$, a trajectory $\mathcal{T}$ is randomly generated within this grid to model the movement of the UT along this path. Then, the channel responses $\mathbf{h}_{{{\mathbf{\Lambda }} _{i,j}}}(t)$ are measured at different time slots $t$ and collected into a set ${\mathcal{H}}= \left\{ \mathbf{h}_{{{\mathbf{\Lambda }} _{i,j}}}(t) \right\}_{t = 1}^T $. Finally, the sCSI at ${{\mathbf{\Lambda }} _{i,j}}$ can be obtained through the calculation of $\mean\left\lbrace \mathbf{h}_{{{\mathbf{\Lambda }} _{i,j}}}(t) \mathbf{h}_{{{\mathbf{\Lambda }} _{i,j}}}(t)^{H} \right\rbrace$. 

Under our system model, the sCSI at the $(i,j)$-th spatial grid is represented by an $N \times N$ covariance matrix $\bm{\Omega}_{{{\mathbf{\Lambda }} _{i,j}}}$, which can be sequentially vectorized into a channel knowledge vector of length $N^2$. For the entire area of interest ${A}$, the vectors at all possible locations are arranged along the spatial dimensions (i.e., the $X$- and $Y$-axes) to jointly form a three-dimensional tensor $\boldsymbol{\mathcal{F}}\in \mathbb{C}^{\sigma \times \sigma \times N^2}$, referred to as the sCF, as illustrated in \figref{fig:1}. It is evident that the number of parameters to be estimated in the sCF is directly determined by the size of the region of interest, the spatial resolution, and the number of BS antennas. In massive MIMO systems, acquiring a high-resolution sCF entails substantial computational, measurement, and sensing overheads, especially in next-generation extremely large-scale MIMO systems.

As analyzed in Subsection \ref{subsection:channel}, the CSCM $\bm{\Omega}_{{{{\mathbf{x}}_k}}}$ can be reconstructed from the CPAS $\widetilde{\bm{\Omega}}_ {{{\mathbf{x}}_k}}={\rm diag} \left\lbrace  \mathcal{S}_{{{{\mathbf{x}}_k}}}\right\rbrace$. Leveraging this relationship, we further define the CPAS as the sCF in this paper, denoted as $\widetilde {\boldsymbol{\mathcal{F}}} \in {\mathbb{R}^{\sigma  \times \sigma  \times N}}$, enabling a dimension reduction from $\sigma  \times \sigma \times N^2$ to $\sigma  \times \sigma \times N$ while effectively preserving the statistical characteristics of the channel. For the structured sCF, the channel knowledge vector at the $(i,j)$-th grid can be expressed as
\begin{align}\label{eq:}
	{{\widetilde {\boldsymbol{\mathcal{F}}}}_{i,j,:}} = {[{\xi _{{{\mathbf{\Lambda }} _{i,j}}}}S_{{{\mathbf{\Lambda }} _{i,j}}}(1),{\xi _{{{\mathbf{\Lambda }} _{i,j}}}}S_{{{\mathbf{\Lambda }} _{i,j}}}(2),...,{\xi _{{{\mathbf{\Lambda }} _{i,j}}}}S_{{{\mathbf{\Lambda }} _{i,j}}}(N)]^T},
\end{align}
where ${{\widetilde {\boldsymbol{\mathcal{F}}}}_{i,j,:}}\in {\mathbb{R}^{N \times 1}}$ denotes all elements along the third dimension at the $(i,j)$-th grid, and are the diagonal elements of CPAS matrix $\widetilde{\bm{\Omega}}_ {{{\mathbf{x}}_k}}$.

\begin{figure}[!b]
	\centering
	\includegraphics[width=0.49\textwidth]{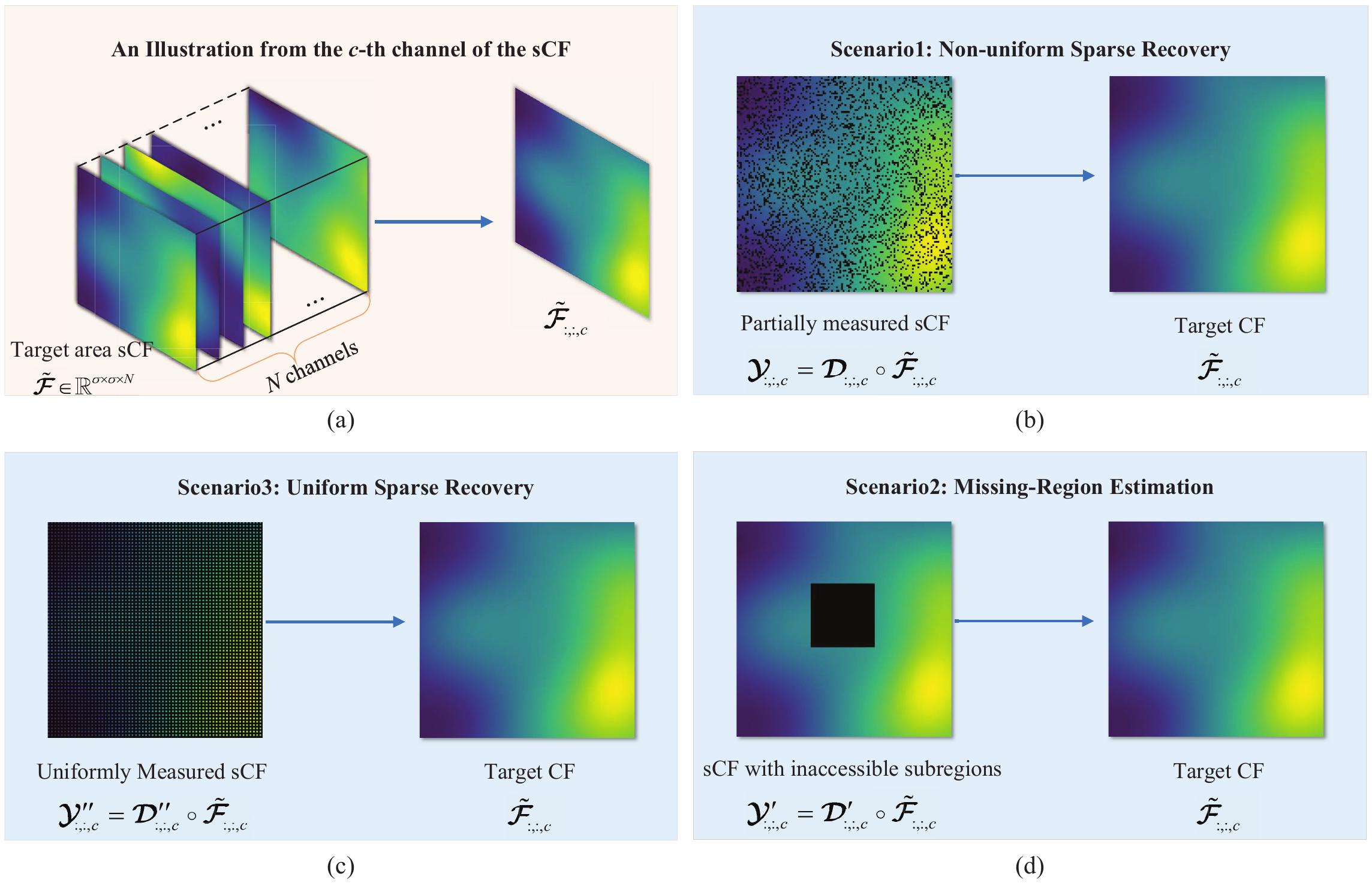}
	\captionsetup{font=footnotesize}
	\caption{Three typical sCF reconstruction tasks under different measurement-constrained scenarios. (a) illustrates the 2D slice of the structured sCF at the $c$-th channel. (b)-(d) depict the reconstruction tasks for the $c$-th sCF channel under three different real-world scenarios. All channels of the sCF are required to undergo these three types of reconstruction tasks.}
	\label{fig:2}
\end{figure}

\subsection{Problem Formulation}\label{sct:pro}
Our objective is to accurately construct the sCF corresponding to the target area. Nevertheless, various practical scenarios may arise due to the high cost of measurements as well as privacy and security restrictions. In real-world scenarios, as depicted in \figref{fig:2}, the construction task of the target sCF can be further categorized into several typical types: \textit{i)} \textit{Non-uniform sparse recovery}: Reconstructing the sCF from partial channel knowledge measured at sparsely and non-uniformly distributed spatial locations; \textit{ii)} \textit{Missing-region estimation}: Estimating the sCF in inaccessible subregions caused by privacy or security constraints; \textit{iii)} \textit{Uniform sparse recovery}: Reconstructing the sCF from partial channel knowledge measured at sparsely yet uniformly distributed spatial locations. 

The task of constructing the target sCF under these distinct practical scenarios can be formulated using a unified degradation model as
\begin{align}\label{eq:12}
	  {\boldsymbol{\mathcal{Y}}}= \mathcal{D}\left( {\widetilde {\boldsymbol{\mathcal{F}}} ;\gamma } \right),
\end{align}
where ${\boldsymbol{\mathcal{Y}}}$ denotes the degraded sCF, i.e., the sCF obtained from practical measurement devices or sensing nodes; $\mathcal{D}( \cdot)$ represents the degradation operator determined by $\gamma$, and $\gamma \in\{-1,0,1\}$ is the degradation task factor, with $\gamma=-1$, $0$, and $1$ corresponding to the non-uniform sparse recovery, the missing-region estimation, and the uniform sparse recovery tasks, respectively. The details of $\mathcal{D}( \cdot)$ corresponding to different scenarios are further discussed as follows.

\textit{i)} When $\gamma=-1$: \eqref{eq:12} can be be rewritten as $ {\boldsymbol{\mathcal{Y}}}={\boldsymbol{\mathcal{D}}}\circ \widetilde {\boldsymbol{\mathcal{F}}}$, where ${\boldsymbol{\mathcal{D}}} \in \mathbb{R}^{\sigma \times \sigma \times N}$ is a randomly generated binary masking tensor. Each element ${\boldsymbol{\mathcal{D}}}_{i',j',k'}$ follows a Bernoulli distribution, i.e., ${\boldsymbol{\mathcal{D}}}_{i',j',k'} \sim {\rm{Bernoulli}}(1-p_{{m}})$, meaning it is retained with probability $1-p_{m}$ and masked to zero with probability $p_{m}$.

\textit{ii)} When $\gamma=0$: Let the inaccessible region be defined as
\begin{align}
	{\mathcal{R}}_{\rm{inaccessible}} \!\!=\!\! \left\{ {(x_{i''},y_{j''})|{x_{\rm{L}}} \leq x_{i''} \leq {x_{\rm{R}}},\;{y_{\rm{B}}} \leq y_{j''} \leq {y_{\rm{T}}}} \right\},
\end{align}
where $x_{\rm{L}}$, $x_{\rm{R}}$, $y_{\rm{B}}$, and $y_{\rm{T}}$ denote the left, right, bottom, and top boundaries of the inaccessible region, respectively. Then, \eqref{eq:12} also can be written as $ {\boldsymbol{\mathcal{Y}}}'={\boldsymbol{\mathcal{D}}}'\circ \widetilde {\boldsymbol{\mathcal{F}}}$, where $\boldsymbol{\mathcal{D}}' \in \mathbb{R}^{\sigma \times \sigma \times N}$ is a deterministic binary mask that encodes a known inaccessible area due to privacy or security constraints. Specifically, the elements of $\boldsymbol{\mathcal{D}}'$ are defined based on the spatial location as
\begin{align}
	{\boldsymbol{\mathcal{D}}}'_{i'',j'',:} = \left\{ \begin{gathered}
		0,\quad{\rm{if}}\,(x_{i''},y_{j''}) \in {\mathcal{R}}_{\rm{inaccessible}}, \hfill \\
		1,\quad{\rm{otherwise}}. \hfill \\ 
	\end{gathered}  \right.
\end{align}

\textit{iii)} When $\gamma=1$: \eqref{eq:12} can be written as ${\boldsymbol{\mathcal{Y}}}'' = {\boldsymbol{\mathcal{D}}}''\circ \widetilde {\boldsymbol{\mathcal{F}}}$, where ${\boldsymbol{\mathcal{D}}}'' \in \mathbb{R}^{\sigma \times \sigma \times N}$ is a uniformly distributed binary masking tensor defined as
\begin{align}
   	{\boldsymbol{\mathcal{D}}}''_{i''',j''',:} = \left\{ \begin{gathered}
   	1,\quad{\rm{if}}\,(i''' \mathrm{mod}\ s =0)\ \mathrm{and}\  (j''' \mathrm{mod} \ s =0) , \hfill \\
   	0,\quad{\rm{otherwise}}, \hfill \\ 
   \end{gathered}  \right.
\end{align}
where $i''',j''' \in \{ {0,\dots,\sigma-1}\}$. This ensures that for a fixed positive integer $s$, every $ls$-th element along both the $X$- and $Y$-directions is retained for any non-negative integer $l$ such that $ls\le\sigma-1$, thereby emulating the uniform sampling intervals in practical measurement processes.

It can be observed that the sCF reconstruction tasks under these different real-world scenarios can all be characterized by the unified formulation in \eqref{eq:12}. These tasks all fall under the category of tensor restoration problems and share a key characteristic, namely inferring the complete sCF over the target area from channel knowledge measured at only a limited number of UT locations, i.e.,
\begin{align}
	\mathcal{M}: {\boldsymbol{\mathcal{Y}}}\to {\widetilde {\boldsymbol{\mathcal{F}}}},
\end{align}
where $\mathcal{M}$ denotes a complicated nonlinear mapping function between the two. As a representative ill-posed inverse problem, it is inherently difficult to derive an explicit analytical solution. Therefore, we are motivated to employ deep learning (DL) to approximate the underlying nonlinear mapping, taking advantage of neural networks' powerful ability in feature extraction and representation. By incorporating carefully designed neural network components, the model can implicitly capture domain-specific priors and spatial correlations of the sCF, enabling more accurate and robust reconstructions even under imperfect measurement conditions. Since the sCF has been modeled as a 3D tensor with an image-like structure (as discussed in Subsection \ref{se:cf}), we further reformulate the sCF construction task as a 3D tensor restoration problem. Consequently, the sCF construction can be represented as
\begin{subequations}\label{eq:}
 \begin{align}
	&\mathop {{\rm{min}}}\limits_\Theta\text{ } \mathcal{L}\left(  {\widetilde {\boldsymbol{\mathcal{F}}}}_m,{{{\hat{\widetilde {\boldsymbol{\mathcal{F}}}}_m}}};\Theta\right)  =\mathbb{E}\left\{ {\left\| {{\widetilde {\boldsymbol{\mathcal{F}}}}_m - {\mathcal{M}}_R\left( {\boldsymbol{\mathcal{Y}}}_m  ;\Theta \right)} \right\|_2^2} \right\},\label{eq:}\\
	&\ \!{\rm{s.t.}}\text{ } {{{{{\mathbf{x}}_k}}}} \in {A} ,m \in \{ 1,2,...,M\},\label{eq:}
 \end{align}
\end{subequations}
where ${\mathcal{M}}_R$ is the proposed model for sCF restoration, ${{{\hat{\widetilde {\boldsymbol{\mathcal{F}}}}_m}}}={\mathcal{M}}_R\left( {\boldsymbol{\mathcal{Y}}}_m  ;\Theta \right)$ denotes the predicted sCF, $\Theta$ represents the set of trainable parameters, ${\boldsymbol{\mathcal{Y}}}_m$ is the observed sCF under constrained measurement conditions, ${\widetilde {\boldsymbol{\mathcal{F}}}}_m$ denotes the ground-truth sCF, and $M$ is the number of sCF samples involved in the training phase.

\section{LPWTNet-Based sCF Construction}\label{sec:LPWTNet-Based sCF Constru}
In this section, we first revisit conventional I2I inpainting methods and analyze two major challenges inherent in existing approaches. Building on these insights, we are then motivated to develop corresponding solutions, with the implementation details of the multi-scale sCF representation via an LP and the small-kernel convolution in the wavelet domain presented step by step. Finally, the operational mechanism and architectural details of the proposed LPWTNet for sCF construction are described.

\subsection{Revisiting Conventional I2I Inpainting Models}\label{sec:rev}
As discussed in Subsection \ref{sct:pro}, the sCF construction task can be formulated as a tensor-to-tensor (T2T) restoration problem, analogous to the widely studied I2I inpainting task in computer vision (CV). This conceptual alignment motivates us to leverage I2I techniques for solving the sCF reconstruction problem. Nevertheless, several key limitations of conventional I2I inpainting models must be addressed before they can be effectively applied to sCF reconstruction tasks.

\textit{i)} Traditional I2I inpainting models are often built upon autoencoder- or U-Net-based architectures \cite{9729564,jin2024i2i,Liang_2021_CVPR}. These models utilize a parameterized encoder-decoder framework where the input is projected into a low-dimensional latent space to separate content and attributes, followed by reconstruction into the original spatial resolution. However, this process generally performs heavy convolution and transposed convolution (deconvolution) operations on the original input, both of which incur considerable computational overhead and increase inference latency. In essence, the reconstruction capability is determined by the learned network parameters of the autoencoder. Consequently, such approaches may be difficult to scale to high-dimensional tensor reconstruction tasks due to their prohibitive computational cost. Furthermore, due to the inherently irreversible nature of these operations, the transformation from high-dimensional input to a low-dimensional latent space and subsequently back to the original dimension inevitably results in the loss of fine-grained structural information \cite{1095851,jin2024i2i}. Meanwhile, U-Net architectures, although capable of retaining certain structural details via skip connections, are still constrained by substantial memory requirements and significant computational burden.

\textit{ii)} It is well known that the size of the receptive field determines a model's ability to capture long-range contextual information and extract high-level features. Given that convolutions in convolutional neural networks (CNNs) inherently perform only local feature mixing, several studies \cite{liu2021swin,liu2022convnet} have explored increasing the convolutional kernel size to approximate the global receptive field achieved by the self-attention blocks in Vision Transformers \cite{dosovitskiy2020image}. Nonetheless, excessively large kernels lead to over-parameterization and introduce substantial computational overhead. Meanwhile, this strategy quickly reaches its limit and saturates before achieving a truly global receptive field. Empirically, performance saturation is typically observed when the kernel size reaches $7\times7$, indicating that further increasing the kernel size yields negligible benefits and, in some cases, may even degrade performance \cite{liu2022convnet}. Moreover, the computational complexity introduced by the self-attention mechanism is also non-negligible.

To address the limitations identified in traditional I2I inpainting models, we propose several tailored architectural modules, each designed to mitigate specific challenges, as elaborated in Subsections \ref{sec: Multi-Scale sCF Representation via Laplacian Pyramid} and \ref{sec: Small Kernel Convolution in the Wavelet Domain}.
\begin{figure}[!b]
	\centering
	\includegraphics[width=0.495\textwidth]{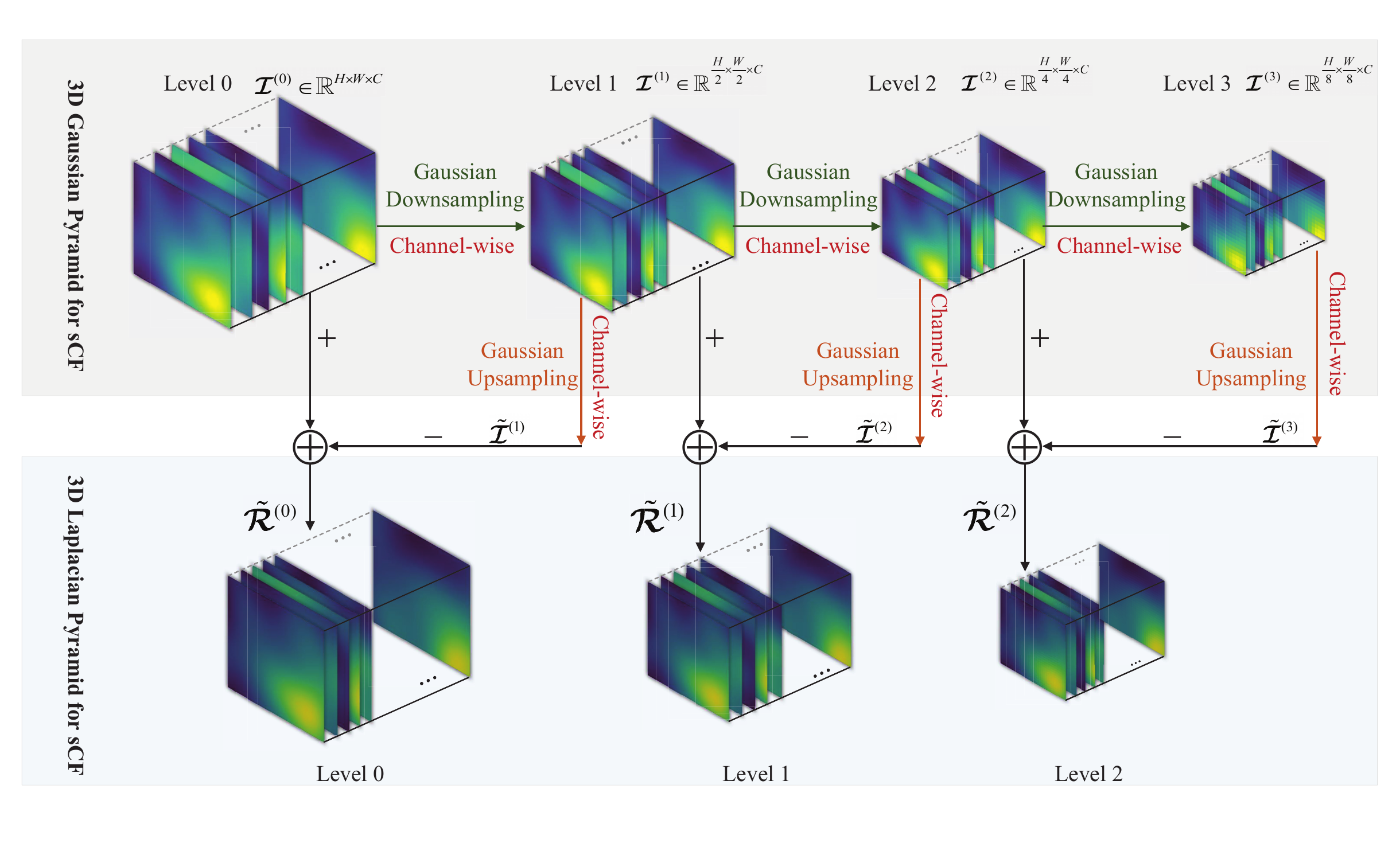}
	\captionsetup{font=footnotesize}
	\caption{Multi-scale high- and low-frequency representations of sCF via 3D LP}
	\label{fig:lpde}
\end{figure}

\subsection{Multi-Scale sCF Representation via Laplacian Pyramid}\label{sec: Multi-Scale sCF Representation via Laplacian Pyramid}
In response to the first limitation discussed in Subsection \ref{sec:rev}, we introduce an LP and its inverse counterpart to replace the conventional parameterized encoder-decoder structure typically utilized in I2I inpainting models. These two components are responsible for input disentanglement and reconstruction, respectively. To mitigate the inherent irreversibility of downsampling and upsampling operations, the LP is employed to retain multi-scale residual information, thereby compensating for the loss of fine-grained details in the reconstruction process. In addition, the LP possesses inherent reversibility and supports closed-form decomposition and reconstruction, allowing for the separation of high- and low-frequency components through fixed filters, thereby eliminating the need for extra trainable parameters \cite{1095851}. Notably, the decomposition process of the LP yields both frequency-separated components and multi-resolution feature maps. Such a combination of multi-frequency and multi-scale representations proves advantageous in various tensor processing tasks \cite{chen2019drop,jin2024i2i,liu2018multi,1095851}.

As shown in \figref{fig:lpde}, the LP is constructed by iteratively applying Gaussian smoothing followed by spatial downsampling to the input feature maps. Let ${\boldsymbol{\mathcal{I}}}^{(0)} \in \mathbb{R}^{H\times W \times C}$ denote the input at level $0$ of the Gaussian pyramid. A low-pass estimation, denoted as ${\boldsymbol{\mathcal{\tilde{I}}}}^{(0)}\in \mathbb{R}^{H\times W \times C}$, is first obtained by independently applying channel-wise 2D Gaussian convolution to each channel of the input feature map:
\begin{align}
	   {\boldsymbol{\mathcal{\tilde{I}}}}^{(0)}_{:,:,c}={\boldsymbol{\mathcal{G}}}_{\rm{2D}}\circledast{\boldsymbol{\mathcal{I}}}^{(0)}_{:,:,c}, \text{   }\forall c=1,...,C, 
\end{align}
where ${\boldsymbol{\mathcal{G}}}_{\rm{2D}}=\mathbf{p}^T\mathbf{p}/256$ represents the 2D Gaussian kernel. The vector $\mathbf{p}=[1,4,6,4,1]$ is a 1D Gaussian filter derived from Pascal's triangle coefficients, serving as a discrete approximation of the Gaussian distribution \cite{1095851}.

Then, $ {\boldsymbol{\mathcal{\tilde{I}}}}^{(0)}$ is spatially downsampled by a factor of 2 to obtain the low-resolution representation for the next level of the pyramid:
\begin{align}
	{\boldsymbol{\mathcal{{I}}}}^{(1)}={\rm{Down}}\left( {\boldsymbol{\mathcal{\tilde{I}}}}^{(0)}\right) \in \mathbb{R}^{\frac{H}{2}\times \frac{W}{2} \times {C}},
\end{align}
where $\text{Down}(\cdot)$ denotes a downsampling operation that retains even indexed elements along both spatial dimensions. To guarantee the reversibility of the reconstruction process, the LP explicitly preserves the high-frequency residual information ${\boldsymbol{\mathcal{\tilde{R}}}}^{(0)}$, calculated as 
\begin{align}
   {\boldsymbol{\mathcal{\tilde{R}}}}^{(0)}={\boldsymbol{\mathcal{I}}}^{(0)}-{\boldsymbol{\mathcal{{\tilde{I}}}}}^{(1)},
\end{align}
where ${\boldsymbol{\mathcal{{\tilde{I}}}}}^{(1)} \in \mathbb{R}^{H\times W \times C}$ represents the result obtained by applying the corresponding upsampling operator followed by Gaussian smoothing to ${\boldsymbol{\mathcal{{I}}}}^{(1)}$. Assuming that the LP has $L$ levels, the aforementioned procedure is performed iteratively on the subsequent feature maps $\{{\boldsymbol{\mathcal{I}}}^{(1)},{\boldsymbol{\mathcal{I}}}^{(2)},...,{\boldsymbol{\mathcal{I}}}^{(L)}\}$, thereby generating a series of residual components $\mathcal{R}=\{{\boldsymbol{\mathcal{\tilde{R}}}}^{(0)},{\boldsymbol{\mathcal{\tilde{R}}}}^{(1)},...,{\boldsymbol{\mathcal{\tilde{R}}}}^{(L-1)}\}$, each corresponding to a specific level of the LP. This process generates a set of multi-scale low- and high-frequency components, which serve as informative representations for the subsequent reconstruction of the sCF.

\subsection{Small Kernel Convolution in the Wavelet Domain}\label{sec: Small Kernel Convolution in the Wavelet Domain}
In response to the second limitation discussed in Subsection \ref{sec:rev}, we introduce a wavelet-domain convolution approach that enables a larger receptive field while using smaller convolution kernels, referred to as WTConv. To be self-contained, we first describe how the wavelet transform is implemented through convolution, and then propose our strategy for small-kernel convolution in the wavelet domain to more efficiently extract sCF feature information.

Without loss of generality, we consider the Haar WT due to its efficiency and broad applicability \cite{finder2022wavelet,liu2018multi}. Note that the proposed approach can also be applied with other wavelet bases. Let ${\boldsymbol{\mathcal{X}}}\in \mathbb{R}^{H\times W\times C}$ denote the input. The one-level Haar WT along a single spatial dimension (either width or height) is achieved through a depth-wise convolution (DWConv) using the kernels $[1,1]/\sqrt 2 $ and $[1,-1]/\sqrt 2 $, followed by a downsampling operation with a factor of 2. To implement the 2D Haar WT, we extend the operation across both spatial dimensions, performing a DWConv with a stride of 2 using the following set of four filters:
\begin{subequations} \label{eq:allfilters}
\begin{align} \label{eq:filters}
	\mathbf{f}_{\mathrm{LL}} &= \frac{1}{2} \begin{bmatrix} 1 & 1 \\ 1 & 1 \end{bmatrix}, \mathbf{f}_{\mathrm{LH}} = \frac{1}{2} \begin{bmatrix} 1 & -1 \\ 1 & -1 \end{bmatrix}, \\
	\mathbf{f}_{\mathrm{HL}} &= \frac{1}{2} \begin{bmatrix} 1 & 1 \\ -1 & -1 \end{bmatrix}, \mathbf{f}_{\mathrm{HH}} = \frac{1}{2} \begin{bmatrix} 1 & -1 \\ -1 & 1 \end{bmatrix},
\end{align}
\end{subequations}
where $\mathbf{f}_{\mathrm{LL}}$represents the low-pass filter, responsible for capturing the low-frequency information in the sCF, and $\mathbf{f}_{\mathrm{LH}}$, $\mathbf{f}_{\mathrm{HL}}$, and $\mathbf{f}_{\mathrm{HH}}$ correspond to the high-pass filters that extract high-frequency information in the horizontal, vertical, and diagonal directions, respectively. For each channel of ${\boldsymbol{\mathcal{X}}}$, the output of the convolution will have four channels, i.e., 
\begin{align}\label{eq:convwt}
	&\left[\left( {{{\boldsymbol{\mathcal{X}}}_{:,:,c}}}\right)_\mathrm{LL},\left( {{{\boldsymbol{\mathcal{X}}}_{:,:,c}}}\right)_\mathrm{LH},\left( {{{\boldsymbol{\mathcal{X}}}_{:,:,c}}}\right)_\mathrm{HL},\left( {{{\boldsymbol{\mathcal{X}}}_{:,:,c}}}\right)_\mathrm{HH} \right] \nonumber \\
	&\quad\quad\quad\quad= \mathrm{DWConv}\left(\left[\mathbf{f}_{\mathrm{LL}}, \mathbf{f}_{\mathrm{LH}}, \mathbf{f}_{\mathrm{HL}}, \mathbf{f}_{\mathrm{HH}}, \right], {\boldsymbol{\mathcal{X}}}_{:,:,c}\right). 
\end{align}
Here, $\left( {{{\boldsymbol{\mathcal{X}}}_{:,:,c}}}\right)_\mathrm{LL}$, $\left( {{{\boldsymbol{\mathcal{X}}}_{:,:,c}}}\right)_\mathrm{LH}$, $\left( {{{\boldsymbol{\mathcal{X}}}_{:,:,c}}}\right)_\mathrm{HL}$, and $\left( {{{\boldsymbol{\mathcal{X}}}_{:,:,c}}}\right)_\mathrm{HH}$ represent the low-frequency component and the horizontal, vertical, and diagonal high-frequency components of $\boldsymbol{\mathcal{X}}{:,:,c}$, respectively, each with half the resolution of $\boldsymbol{\mathcal{X}}{:,:,c}$. Since the filters in \eqref{eq:allfilters} mathematically form an orthogonal basis, the inverse wavelet transform (IWT) can be performed using transposed convolutions to merge the frequency components and reconstruct the original input:
\begin{align}
	\!\!\!\boldsymbol{\mathcal{X}}_{:,:,c} &= \mathrm{DWConv}^\mathrm{Transposed}\left(\left[\mathbf{f}_{\mathrm{LL}},\mathbf{f}_{\mathrm{LH}},\mathbf{f}_{\mathrm{HL}},\mathbf{f}_{\mathrm{HH}} \right], \right.\nonumber \\
	&\left. \left[\left( {{{\boldsymbol{\mathcal{X}}}_{:,:,c}}}\right)_\mathrm{LL}, \left( {{{\boldsymbol{\mathcal{X}}}_{:,:,c}}}\right)_\mathrm{LH}, \left( {{{\boldsymbol{\mathcal{X}}}_{:,:,c}}}\right)_\mathrm{HL}, \left( {{{\boldsymbol{\mathcal{X}}}_{:,:,c}}}\right)_\mathrm{HH} \right]   \right).
\end{align}
To realize cascaded wavelet decomposition, the WT is recursively applied to the low-frequency component, and the decomposition at each level can be expressed as
\begin{align}\label{eq:cascaded}
	\left[ \boldsymbol{\mathcal{X}}_\mathrm{LL}^{(i)}, \boldsymbol{\mathcal{X}}_\mathrm{LH}^{(i)}, \boldsymbol{\mathcal{X}}_\mathrm{HL}^{(i)}, \boldsymbol{\mathcal{X}}_\mathrm{HH}^{(i)}\right] =\mathrm{WT}\left( \boldsymbol{\mathcal{X}}_\mathrm{LL}^{(i-1)}\right),
\end{align}
where $\boldsymbol{\mathcal{X}}_\mathrm{LL}^{(0)}=\boldsymbol{\mathcal{X}}$, with $i$ indicating the current decomposition level. As $i$ increases, the frequency resolution of the low-frequency component improves, while its spatial resolution correspondingly decreases.
\begin{figure}[!t]
	\centering
	\includegraphics[width=0.49\textwidth]{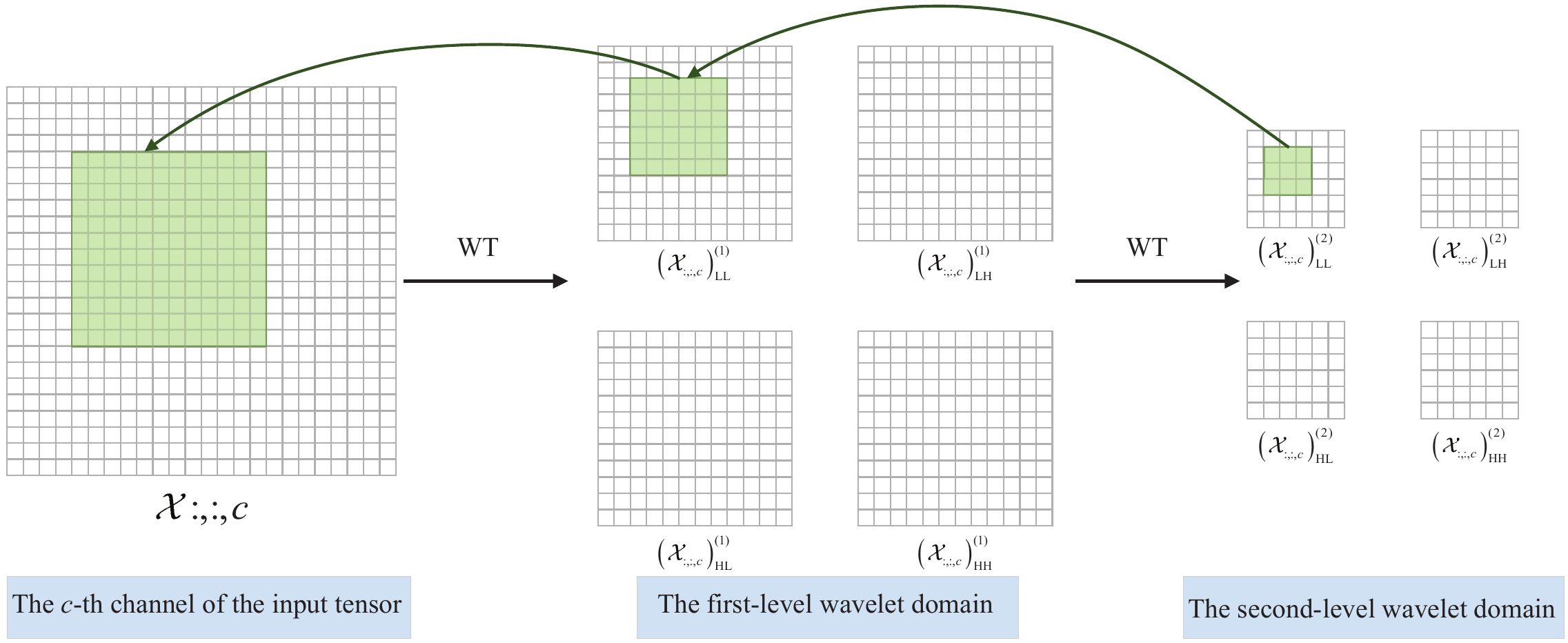}
	\captionsetup{font=footnotesize}
	\caption{Illustration of performing small-kernel DWConv in the wavelet domain. In this case, a $3 \times 3$ convolution kernel applied to the second-level wavelet domain $\left( {{{\boldsymbol{\mathcal{X}}}_{:,:,c}}}\right)^{(2)}_\mathrm{LL}$ corresponds to a $12 \times 12$ receptive field in the input $\left( {{{\boldsymbol{\mathcal{X}}}_{:,:,c}}}\right)$, primarily capturing lower-frequency components.}
	\label{fig:wtconv}
\end{figure}

In conventional convolution operation, the receptive field is typically enlarged by increasing the kernel size. However, this approach incurs a quadratic order growth in the number of parameters, thereby leading to considerable computational overhead. To address this limitation, we introduce WTConv, which integrates WT with small DWConv kernels. Specifically, the WT first filters the input and then downsamples both the low- and high-frequency components. Subsequently, compact DWConv kernels are applied to these frequency-band feature maps, followed by the IWT to construct the output. In this manner, convolution can be performed over a larger receptive field while maintaining computational efficiency. As illustrated in \figref{fig:wtconv}, this process not only decouples the convolution across different frequency components but also enables a smaller kernel to operate over a larger region of the original input, thereby effectively increasing the receptive field with respect to the input. Based on the cascaded decomposition principle \eqref{eq:cascaded}, this process can be further expressed as
\begin{subequations} 
	\begin{align} \label{eq:}
		\left[ \boldsymbol{\mathcal{X}}_\mathrm{LL}^{(i)},  \boldsymbol{\mathcal{X}}_\mathrm{H}^{(i)}\right] &=\mathrm{WT}\left( \boldsymbol{\mathcal{X}}_\mathrm{LL}^{(i-1)}\right), \\
		\left[ \boldsymbol{\mathcal{Z}}_\mathrm{LL}^{(i)},  \boldsymbol{\mathcal{Z}}_\mathrm{H}^{(i)}\right] &=\mathrm{DWConv}_{k\times k}\left(\boldsymbol{\mathcal{W}}^{(i)},\left( \boldsymbol{\mathcal{X}}_\mathrm{LL}^{(i)},  \boldsymbol{\mathcal{X}}_\mathrm{H}^{(i)}\right) \right),
	\end{align}
\end{subequations}
where $\boldsymbol{\mathcal{X}}_\mathrm{LL}^{(i)}$ and $\boldsymbol{\mathcal{X}}_\mathrm{H}^{(i)}=[ \boldsymbol{\mathcal{X}}_\mathrm{LH}^{(i)},\boldsymbol{\mathcal{X}}_\mathrm{HL}^{(i)},\boldsymbol{\mathcal{X}}_\mathrm{HH}^{(i)}]$ represent the low-frequency and high-frequency component sets at the $i$-th level, respectively. $\boldsymbol{\mathcal{W}}^{(i)}$ denotes the learnable $k \times k$ DWConv kernel during training, where the padding size is set to $(k-1)/2$ and the stride is fixed to 1 to ensure that the output feature map has the same spatial dimensions as the input. $\boldsymbol{\mathcal{Z}}_\mathrm{LL}^{(i)}$ and $\boldsymbol{\mathcal{Z}}_\mathrm{H}^{(i)}=[ \boldsymbol{\mathcal{Z}}_\mathrm{LH}^{(i)},\boldsymbol{\mathcal{Z}}_\mathrm{HL}^{(i)},\boldsymbol{\mathcal{Z}}_\mathrm{HH}^{(i)}]$ represent the outputs obtained by applying DWConv to the low-frequency and high-frequency component sets at the same $i$-th level. At each level, the WT progressively refines the low-frequency components of the input. Moreover, to enable the model to dynamically adjust the relative importance of frequency components across different levels during feature learning, we introduce a wavelet scale layer after each stage of wavelet decomposition. This layer is applied to both the low- and high-frequency subbands to adaptively modulate their contributions to the final output. The wavelet scaling process is given by
\begin{align}
	\left[ \boldsymbol{\mathcal{\tilde Z}}_\mathrm{LL}^{(i)},  \boldsymbol{\mathcal{\tilde Z}}_\mathrm{H}^{(i)}\right] = \zeta^{(i)} \left[ \boldsymbol{\mathcal{ Z}}_\mathrm{LL}^{(i)},  \boldsymbol{\mathcal{ Z}}_\mathrm{H}^{(i)}\right],
\end{align}
where $\zeta^{(i)}$ is a learnable scaling factor that enables the network to progressively determine the importance of each frequency subband during training, thereby allowing different frequency components to contribute in a more fine-grained and adaptive manner. Leveraging the fact that both the WT and its inverse are linear operations, multiple frequency components from different levels can be integrated to reconstruct the output tensor, i.e.,
\begin{align}
	\boldsymbol{\mathcal{\hat Z}}^{(i)}=\mathrm{IWT}\left(\left( \boldsymbol{\mathcal{\tilde Z}}_\mathrm{LL}^{(i)}+\boldsymbol{\mathcal{\hat Z}}^{(i+1)}\right) ,\boldsymbol{\mathcal{\tilde Z}}_\mathrm{H}^{(i)}\right) ,
\end{align}
where $\boldsymbol{\mathcal{\hat Z}}^{(i)}$ represents the aggregated outputs starting from level $i$. For clarity, the implementation of WTConv with $k\times k$ DWConv kernels is summarized in \textbf{\alref{alg:infer}}.
\begin{algorithm}[!t]
	\caption{$\mathrm{WTConv}_{k\times k}$ Implementation Procedure}
	\label{alg:infer}
	\begin{algorithmic}[1]
		\STATE
		\textbf{Input:} ${\boldsymbol{\mathcal{X}}}\in \mathbb{R}^{H\times W\times C}$, $\boldsymbol{\mathcal{X}}_\mathrm{LL}^{(0)}={\boldsymbol{\mathcal{X}}}$, $ \boldsymbol{\mathcal{Z}}_\mathrm{LL}^{(0)} =\mathrm{DWConv}_{k\times k}\left(\boldsymbol{\mathcal{W}}^{(0)}, \boldsymbol{\mathcal{X}}_\mathrm{LL}^{(0)}\right)$, $ \boldsymbol{\mathcal{\tilde Z}}_\mathrm{LL}^{(0)} = \zeta^{(0)}\boldsymbol{\mathcal{Z}}_\mathrm{LL}^{(0)}$
		\FOR{$i=1,..,\ell-1$}
		\vspace{0.1cm}
		\STATE
		$[ \boldsymbol{\mathcal{X}}_\mathrm{LL}^{(i)},  \boldsymbol{\mathcal{X}}_\mathrm{H}^{(i)}] =\mathrm{WT}\left( \boldsymbol{\mathcal{X}}_\mathrm{LL}^{(i-1)}\right)$
		\STATE
		$[ \boldsymbol{\mathcal{Z}}_\mathrm{LL}^{(i)},  \boldsymbol{\mathcal{Z}}_\mathrm{H}^{(i)}] =\mathrm{DWConv}_{k\times k}\left(\boldsymbol{\mathcal{W}}^{(i)},\left( \boldsymbol{\mathcal{X}}_\mathrm{LL}^{(i)},  \boldsymbol{\mathcal{X}}_\mathrm{H}^{(i)}\right) \right)$
		\STATE
		$[ \boldsymbol{\mathcal{\tilde Z}}_\mathrm{LL}^{(i)},  \boldsymbol{\mathcal{\tilde Z}}_\mathrm{H}^{(i)}] = \zeta^{(i)} \left[ \boldsymbol{\mathcal{ Z}}_\mathrm{LL}^{(i)},  \boldsymbol{\mathcal{ Z}}_\mathrm{H}^{(i)}\right]$
		\ENDFOR
		\vspace{0.12cm}
		\STATE
		$\boldsymbol{\mathcal{\hat Z}}^{(\ell)}=0$
		\FOR{$i=\ell-1$,..,1}
		\vspace{0.1cm}
		\STATE
		$\boldsymbol{\mathcal{\hat Z}}^{(i)}=\mathrm{IWT}\left(\left( \boldsymbol{\mathcal{\tilde Z}}_\mathrm{LL}^{(i)}+\boldsymbol{\mathcal{\hat Z}}^{(i+1)}\right) ,\boldsymbol{\mathcal{\tilde Z}}_\mathrm{LH}^{(i)},\boldsymbol{\mathcal{\tilde Z}}_\mathrm{HL}^{(i)},\boldsymbol{\mathcal{\tilde Z}}_\mathrm{HH}^{(i)}\right)$
		\ENDFOR
		\vspace{0.12cm}
		\STATE
		$\boldsymbol{\mathcal{\hat Z}}^{(0)}= \boldsymbol{\mathcal{\tilde Z}}_\mathrm{LL}^{(0)} + \boldsymbol{\mathcal{\hat Z}}^{(1)}$
		\RETURN 
		The output of WTConv, $\boldsymbol{\mathcal{\hat Z}}^{(0)}$
	\end{algorithmic}
\end{algorithm} 

\subsection{Advantages of WTConv and Its Complexity Analysis}
There are several key advantages to performing depth-wise convolution in the wavelet domain. First, each level of the WT significantly enlarges the receptive field, while introducing only a marginal increase in the number of trainable parameters. Specifically, for an $\ell$-level WT with a $k\times k$ DWConv applied at each level, the number of trainable parameters increases linearly with the number of levels, proportional to $4Ck^2\ell$, whereas the receptive field expands exponentially with the number of levels, proportional to $2^{\ell}k$. Second, the iterative decomposition process of the WT enhances the representation of low-frequency information, enabling the network to better capture the primary structures and global patterns of the input. This, in turn, compensates for the limitation of standard convolutional layers, which tend to be more sensitive to high-frequency components.

Furthermore, we provide a complexity analysis in terms of floating-point operations (FLOPs). For DWConv, the computational complexity is $\mathcal{O}(CH_{\mathrm{out}}W_{\mathrm{out}}{k^2})$, where $H_{\mathrm{out}}=(H+2P-k)/S+1$ and $W_{\mathrm{out}}=(W+2P-k)/S+1$ denote the height and width of the output feature map, respectively, $P$ is the padding size, and $S$ is the stride. For the proposed WTConv, each wavelet-domain convolution is performed on spatial dimensions reduced by a factor of 2, while the number of channels is four times that of the original input. Its computational complexity is $\mathcal{O}(C{k^2}(HW+\sum\nolimits_{i=1}^{\ell} {4HW/2^{2i}}))$. Note that the computational cost of WT itself should also be taken into account. When the Haar basis is used, WTs can be efficiently implemented via standard DWConv operations as defined in \eqref{eq:allfilters}-\eqref{eq:convwt}, with a computational complexity of $\mathcal{O}(4C\sum\nolimits_{i=0}^{\ell-1} {HW/2^{2i}})$. Similarly, the computational complexity of the IWT is identical to that of the WT. For clarity, let us consider an example with a single-channel input of spatial dimension $256\times256$. Performing an $11\times11$ standard convolution with ``same'' padding results in approximately 7.9 M FLOPs. Increasing the kernel size from $11\times11$ to $31\times31$ to enlarge the receptive field raises the computational cost to 63.0 M FLOPs. In contrast, for the proposed WTConv, performing the multi-frequency convolutions of a 3-level WTConv with a kernel $\boldsymbol{\mathcal{W}}$ of size $5\times5$ yields an effective receptive field of $40\times40=(5\cdot2^3)\times(5\cdot2^3)$ while requiring only 3.8 M FLOPs. Moreover, the 3-level WT and IWT introduce an additional computational cost of 0.7 M FLOPs. Therefore, in this example, the overall computational cost of the proposed WTConv is 4.4 M FLOPs, representing a substantial reduction compared to that of a standard DWConv achieving the same receptive field size.

\subsection{LPWTNet-Based sCF Construction Network}
\begin{figure*}[!t]
	\centering
	\includegraphics[width=0.95\textwidth]{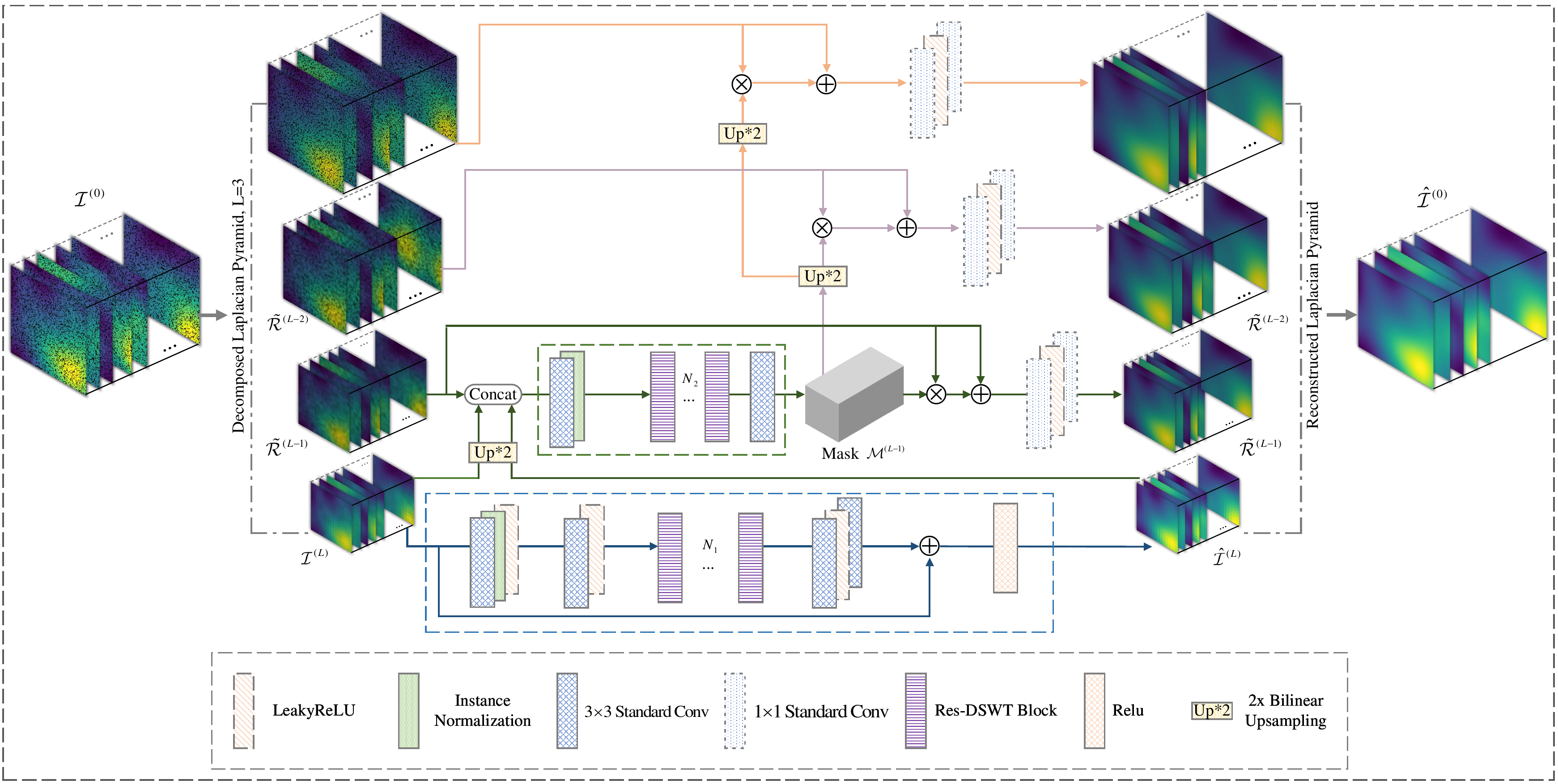}
	\captionsetup{font=footnotesize}
	\caption{Illustration of the workflow for the proposed LPWTNet. ${\boldsymbol{\mathcal{I}}}^{(0)} \in \mathbb{R}^{H \times W \times C}$ and ${\boldsymbol{\hat{\mathcal{I}}}}^{(0)} \in \mathbb{R}^{H \times W \times C}$ have dimensions of $32 \times 32 \times 64$ in the experiments described in Section \ref{sec:Numerical Experiment}.}
	\label{fig:lpnet}
\end{figure*}

To address the limitations discussed in Subsection \ref{sec:rev}, we propose an end-to-end architecture, referred to as the LPWT-based sCF construction network (LPWTNet), which expands the receptive field while reducing computational cost to facilitate efficient sCF construction. The overall pipeline of the proposed LPWTNet is illustrated in \figref{fig:lpnet}.

As shown in \figref{fig:lpnet}, given an sCF ${\boldsymbol{\mathcal{I}}}^{(0)}\in {\mathbb{R}^{H \times W \times C}}$, we first decompose it utilizing an $L$-level LP, obtaining a set of high-frequency residual components $\mathcal{R}=\{{\boldsymbol{\mathcal{\tilde{R}}}}^{(0)},{\boldsymbol{\mathcal{\tilde{R}}}}^{(1)},...,{\boldsymbol{\mathcal{\tilde{R}}}}^{(L-1)}\}$, together with a low-frequency component ${{\boldsymbol{\mathcal{I}}}^{(L)}} \in {\mathbb{R}^{\tfrac{H}{{{2^L}}} \times \tfrac{W}{{{2^L}}} \times C}}$. The spatial resolutions of the components in $\mathcal{R}$ progressively decrease from $H\times W$ to $\tfrac{H}{{{2^L}}} \times \tfrac{W}{{{2^L}}}$. For the low-frequency sCF components, which convey more global structural information \cite{1095851}, we construct the primary representation based on ${{\boldsymbol{\mathcal{I}}}^{(L)}}$. Meanwhile, the high-frequency component set $\mathcal{R}$ is progressively and adaptively refined under the guidance of the lower-frequency components. Specifically, the proposed LPWTNet consists of three main components. First, we transform ${{\boldsymbol{\mathcal{I}}}^{(L)}}$ into ${{\boldsymbol{\mathcal{\hat I}}}^{(L)}}$ through the proposed deep reconstruction network. Second, we learn a mask based on the concatenation of $[{\boldsymbol{\mathcal{\tilde{R}}}}^{(L-1)}, \mathrm{up}({{\boldsymbol{\mathcal{I}}}_L}),\mathrm{up}({{\boldsymbol{\mathcal{\hat I}}}_L})]$, where $\mathrm{up}(\cdot)$ denotes a bilinear upsampling operation. This mask is then multiplied with ${\boldsymbol{\mathcal{\tilde{R}}}}^{(L-1)}$ to refine the high-frequency components at the $(L-1)$-th level. Finally, to progressively refine the remaining higher-frequency components, we introduce a shared mask learning strategy. For each level from $l=L-2$ to $l=0$, the mask obtained from the previous level is first upsampled and then lightly fine-tuned using a lightweight convolutional module. The overall framework is detailed below.
\begin{figure}[!t]
	\centering
	\includegraphics[width=0.45\textwidth]{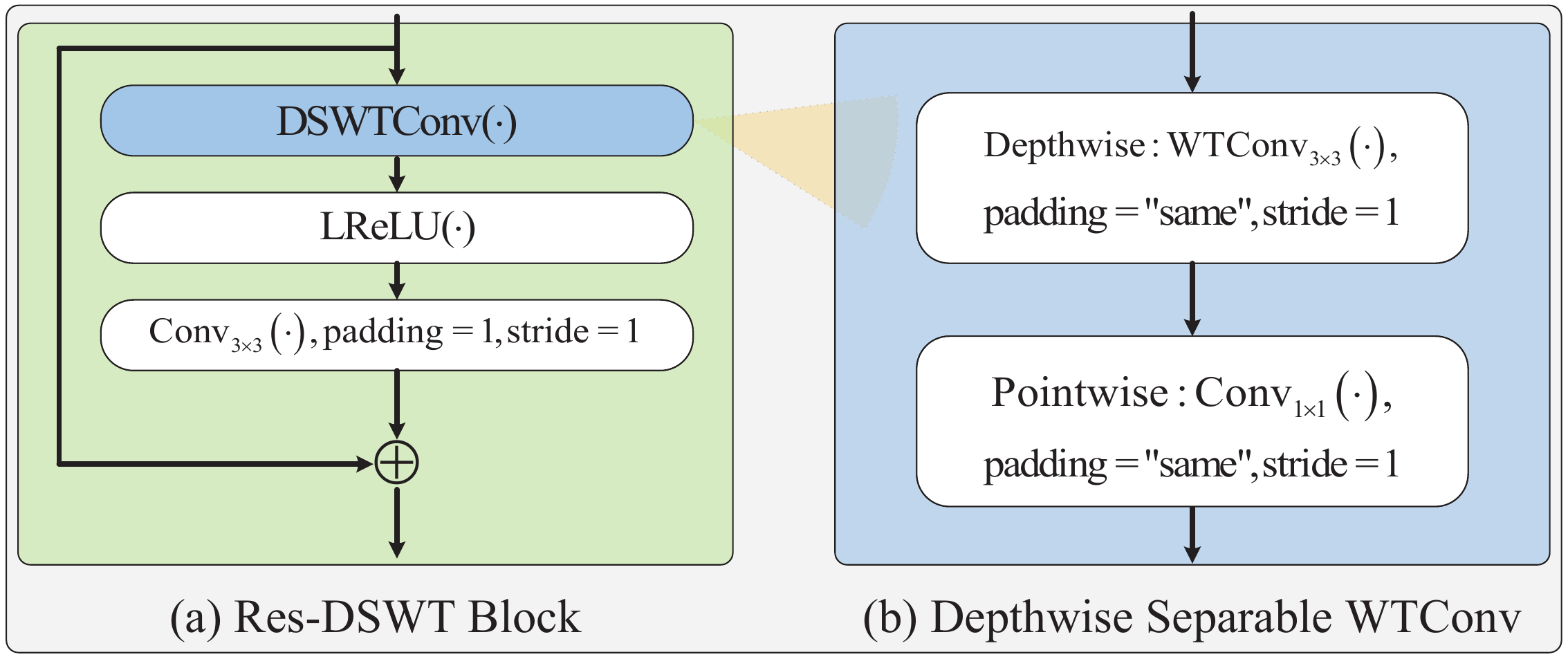}
	\captionsetup{font=footnotesize}
	\caption{Structural diagram of the proposed Res-DSWT module.}
	\label{fig:resdpwtconv}
\end{figure}

\textit{i) Construction on Low-Frequency sCF:}
As shown in \figref{fig:lpnet}, given the low-frequency sCF ${{\boldsymbol{\mathcal{I}}}^{(L)}} \in {\mathbb{R}^{\tfrac{H}{{{2^L}}} \times \tfrac{W}{{{2^L}}} \times C}}$ with reduced spatial resolution, we first apply two standard $3\times3$ convolutional layers with the padding set to 1 and the stride set to 1 to increase the channel dimension of the feature map, i.e., 
\begin{align}\label{eq:}
	{{\boldsymbol{\mathcal{I'}}}^{(L)}} \!\!=\! \mathrm{LReLU}\!\left(\!\mathrm{Conv}_{3\times3}\!\left(\!\mathrm{LReLU}\!\left(\! \mathrm{IN}\!\left(\! \mathrm{Conv}_{3\times3}\!\left(\! {{\boldsymbol{\mathcal{I}}}^{(L)}}\!\right)\! \right)\! \right)\! \right)\! \right),
\end{align}
where $\mathrm{IN}(\cdot)$ denotes an instance normalization layer and $\mathrm{LReLU}(\cdot)$ denotes a LeakyReLU activation layer. Next, we stack $N_1$ of the proposed residual depthwise separable WT convolution (Res-DSWT) blocks on the expanded feature maps to further extract features in an efficient manner. As shown in \figref{fig:resdpwtconv}, the proposed Res-DSWT block adopts depthwise separable WTConv to replace the standard convolution. Specifically, the output of the Res-DSWT block is given by
\begin{align}\label{eq:}
	{{\boldsymbol{\mathcal{I''}}}^{(L)}} \!=\!\mathrm{Conv}_{3\times3}\!\left(\!\mathrm{LReLU}\!\left(\! \mathrm{DSWTConv}\!\left(\! {{\boldsymbol{\mathcal{I'}}}^{(L)}} \!\right)\! \right)\! \right)\!+\!{{\boldsymbol{\mathcal{I'}}}^{(L)}},
\end{align}
where $\mathrm{DSWTConv}(\cdot)=\mathrm{Conv}_{1\times1}( \mathrm{WTConv}_{3\times3}( \cdot))$ denotes the depthwise separable convolution implemented with WTConv. Subsequently, we apply two standard $3\times3$ convolutional layers followed by a LeakyReLU activation layer to reduce the channel dimension of the feature map back to that of the original input, $C$, expressed as 
\begin{align}\label{eq:}
	{{\boldsymbol{\mathcal{I'''}}}^{(L)}} =\mathrm{Conv}_{3\times3}\left(\mathrm{LReLU}\left( \mathrm{Conv}_{3\times3}\left( {{\boldsymbol{\mathcal{I''}}}^{(L)}} \right) \right) \right).
\end{align}
Finally, ${{\boldsymbol{\mathcal{I'''}}}^{(L)}}$ is added to the original low-frequency sCF, followed by a ReLU activation layer, to obtain the reconstructed result on the low-frequency sCF, denoted as ${{\boldsymbol{\mathcal{\hat I}}}^{(L)}}$, i.e.,
\begin{align}\label{eq:}
{{\boldsymbol{\mathcal{\hat I}}}^{(L)}} =\mathrm{ReLU}\left({{\boldsymbol{\mathcal{I}}}^{(L)}}+{{\boldsymbol{\mathcal{I'''}}}^{(L)}}\right) .
\end{align}

\textit{ii) Refinement on High-Frequency sCFs:} The low-frequency component reflects the global structural attributes of the input and provides strong global priors \cite{1095851}. Therefore, the high-frequency components $\mathcal{R}=\{{\boldsymbol{\mathcal{\tilde{R}}}}^{(0)},{\boldsymbol{\mathcal{\tilde{R}}}}^{(1)},...,{\boldsymbol{\mathcal{\tilde{R}}}}^{(L-1)}\}$, which correspond to local details, should be refined through dynamic adjustments guided by the low-frequency component. To this end, we propose to learn a mask for the high-frequency sCF ${\boldsymbol{\mathcal{\tilde{R}}}}^{(L-1)}$ and progressively expand this mask to refine the remaining higher-frequency components $\mathcal{R}=\{{\boldsymbol{\mathcal{\tilde{R}}}}^{(0)},{\boldsymbol{\mathcal{\tilde{R}}}}^{(1)},...,{\boldsymbol{\mathcal{\tilde{R}}}}^{(L-2)}\}$, according to the intrinsic decomposition mechanism of the LP. First, we upsample ${{\boldsymbol{\mathcal{\hat I}}}^{(L)}} \in {\mathbb{R}^{\tfrac{H}{{{2^L}}} \times \tfrac{W}{{{2^L}}} \times C}}$ and ${{\boldsymbol{\mathcal{I}}}^{(L)}}\in {\mathbb{R}^{\tfrac{H}{{{2^L}}} \times \tfrac{W}{{{2^L}}} \times C}}$ using bilinear operations to resize their spatial dimensions to ${\textstyle{H \over {{2^{L - 1}}}}} \times {\textstyle{W \over {{2^{L - 1}}}}}$, so as to match the resolution of the high-frequency sCF ${\boldsymbol{\mathcal{\tilde{R}}}}^{(L-1)}\in {\mathbb{R}^{\tfrac{H}{{{2^{L-1}}}} \times \tfrac{W}{{{2^{L-1}}}} \times C}}$. Then, we concatenate $[{{\boldsymbol{\mathcal{\hat I}}}^{(L)}}, {{\boldsymbol{\mathcal{I}}}^{(L)}}, {\boldsymbol{\mathcal{\tilde{R}}}}^{(L-1)}]$ along the channel dimension to form ${{\boldsymbol{\mathcal{F}}}^{(L-1)}}$, and feed it into a lightweight network that adopts the same architecture as shown in \figref{fig:lpnet}. Specifically, this lightweight network consists of one $3\times3$ convolutional layer for expanding the channel dimension, followed by a stack of $N_2$ Res-DSWT blocks for further feature extraction and representation enhancement. Finally, one $3\times3$ convolutional layer is applied to reduce the channel dimension back to $C$, yielding the mask ${{\boldsymbol{\mathcal{M}}}^{(L-1)}}\in {\mathbb{R}^{\tfrac{H}{{{2^{L-1}}}} \times \tfrac{W}{{{2^{L-1}}}} \times C}}$. The mask ${{\boldsymbol{\mathcal{M}}}^{(L-1)}}$ provides a global adjustment mechanism, which is relatively easier to optimize compared with the mixed-frequency input. To ensure that high-frequency details are restored in a structurally consistent manner, we adopt this shared mask to guide the progressive refinement of the remaining high-frequency components. Accordingly, ${\boldsymbol{\mathcal{\tilde{R}}}}^{(L-1)}$ is refined through element-wise multiplication and residual addition, i.e.,
\begin{align}\label{eq:32}
	{\boldsymbol{\mathcal{\hat{R}}}}^{(L-1)}={\boldsymbol{\mathcal{\tilde{R}}}}^{(L-1)}\circ{{\boldsymbol{\mathcal{M}}}^{(L-1)}}+{\boldsymbol{\mathcal{\tilde{R}}}}^{(L-1)}.
\end{align}
Then, a lightweight convolutional block, consisting of two $1\times1$ convolutional layers and a LeakyReLU activation layer, is applied for fine-tuning.

Similarly, we progressively upsample the mask ${{\boldsymbol{\mathcal{M}}}^{(L-1)}}$ to obtain the corresponding mask set $\lbrace {{\boldsymbol{\mathcal{M}}}^{(L-2)}},...,{{\boldsymbol{\mathcal{M}}}^{(1)}},{{\boldsymbol{\mathcal{M}}}^{(0)}}\rbrace $, whose spatial resolutions range from $\tfrac{H}{{{2^{L-2}}}} \times \tfrac{W}{{{2^{L-2}}}} \times C$ to $H\times W \times C$, so as to match the sCFs at different frequency levels. Consequently, the remaining high-frequency sCFs can be gradually refined according to \eqref{eq:32}, yielding the corresponding set $\lbrace 	{\boldsymbol{\mathcal{\hat{R}}}}^{(L-2)},...,	{\boldsymbol{\mathcal{\hat{R}}}}^{(1)},	{\boldsymbol{\mathcal{\hat{R}}}}^{(0)}\rbrace $. Finally, the reconstructed sCF ${{\boldsymbol{\mathcal{\hat I}}}^{(0)}}$ is obtained by combining the reconstructed low-frequency component ${{\boldsymbol{\mathcal{\hat I}}}^{(L)}}$ with the refined high-frequency components $\lbrace 	{\boldsymbol{\mathcal{\hat{R}}}}^{(L-1)},...,	{\boldsymbol{\mathcal{\hat{R}}}}^{(1)},	{\boldsymbol{\mathcal{\hat{R}}}}^{(0)}\rbrace $, leveraging the inverse LP property.

\section{Numerical Experiment}\label{sec:Numerical Experiment}
In this section, we first present the implementation details, including the wireless communication scenario setup, sCF data generation, training strategy, model configuration, and evaluation criteria. Next, ablation experiments are conducted to assess the impact of the proposed DSWTConv on the reconstruction accuracy and convergence performance of the model, as well as to compare it with state-of-the-art (SOTA) modules. Finally, a comprehensive evaluation is performed to compare the proposed approach with the baselines in terms of reconstruction accuracy and computational complexity across three typical sCF construction scenarios.

\subsection{Implementation Details}
\textit{i) Wireless Communication Scenario Setup:} Consider a massive MIMO system operating within an interest communication area $A$, measuring 32 m $\times$ 32 m, where the BS is equipped with an $8\times8$ UPA, with half-wavelength spacing between antennas. The ground-truth channels for \eqref{eq:1} are generated utilizing the widely adopted QuaDRiGa generator (version 2.6.1) \cite{jaeckel2014quadriga}. The channel in \eqref{eq:1} represents a typical MIMO multipath channel \cite{liu2023structured}, with its components and associated parameters, such as channel path gains, elevation angles, azimuth angles, and other relevant factors, generated by QuaDRiGa. Furthermore, we consider the 5G NR typical urban micro-cell scenario ``3GPP 38.901 UMa NLOS'', which includes both LOS and NLOS physical propagation environments. This scenario is representative of the complicated urban environments encountered in practical communication systems. The detailed simulation parameters are summarized in \tabref{ta:sys}.

\textit{ii) sCF Data Generation:} For the generation of sCF data, we distinguish between the target sCF and the three types of imperfect sCF introduced in Subsection \ref{sct:pro}. The former corresponds to the ground-truth and ideal sCF, while the latter simulates the practical sCF obtained in realistic communication scenarios under constraints such as measurement cost. In our simulations, the three imperfect sCFs are derived from the ground-truth sCF through different degradation processes.

Within the aforementioned communication scenario, ${\sigma}$ is set to 32, and a sampling interval of $\Delta_x=\Delta_y=1$ m is employed to discretize the target area $A$. For each grid ${{\mathbf{\Lambda }} _{i,j}}$, the UT trajectory $\mathcal{T}$ is modeled as a circular path for measurement, with 64 time slots sampled along the trajectory, as described in Subsection \ref{se:cf}. Subsequently, for each grid point, we generate the channel set $\left\{ \mathbf{h}_{{{\mathbf{\Lambda }} _{i,j}}}(t) \right\}_{t = 1}^{64}$ and calculate the corresponding CSCM $\bm{\Omega}_{{{\mathbf{\Lambda }} _{i,j}}}$ according to \eqref{eq:r}. Then, we perform eigenvalue decomposition on $\bm{\Omega}_{{{\mathbf{\Lambda }} _{i,j}}}$, and the resulting eigenvalues are arranged into a channel knowledge vector of length $64$. Finally, by sequentially concatenating the channel knowledge vectors across the spatial dimension within the region of interest $A$, a three-dimensional sCF tensor $\widetilde {\boldsymbol{\mathcal{F}}}_m \in {\mathbb{R}^{32  \times 32  \times 64}}$ is constructed, as described in Subsection \ref{se:cf}. Similarly, $M=10,000$ ground-truth sCF samples $\{ \widetilde {\boldsymbol{\mathcal{F}}}_m \}_{m = 1}^{10,000}$ are generated through QuaDRiGa simulations by varying the BS locations $(x,y)$, where $x$ and $y$ are independently drawn from a uniform distribution over $[0,32]$. Additionally, to improve the efficiency of the training process, min-max normalization is applied to the raw sCF \cite{11220249,jin2024i2i}. Accordingly, to emulate the three real-world sCF construction scenarios, namely non-uniform sparse recovery, missing-region estimation and uniform sparse recovery, we adopt the three degradation procedures described in Subsection \ref{sct:pro} to generate the corresponding degraded sCFs, which represent imperfect sCFs encountered in practical measurements. Each degraded sCF is paired with its ground truth counterpart to form datasets corresponding to the three scenarios. The parameter settings for the three degradation processes are summarized in \tabref{ta:sys}. The resulting sCF samples are subsequently divided into training and testing sets at a 4:1 ratio for each considered scenario.
\newcolumntype{L}{>{\hspace*{-\tabcolsep}}l}
\newcolumntype{R}{c<{\hspace*{-\tabcolsep}}}
\definecolor{lightblue}{rgb}{0.93,0.95,1.0}
\begin{table}[!b]
	\captionsetup{font=footnotesize}
	\caption{Wireless Communication System and Model Configuration Parameters}\label{ta:sys}
	\centering
	\setlength{\tabcolsep}{11mm}
	\ra{2.0}
	\scriptsize
	\scalebox{0.8}{\begin{tabular}{LR}
			\toprule
			Parameter &  Value\\
			\midrule
			\rowcolor{lightblue}
			Size of the interested area $A$ & 32 m$\times$32 m\\
			Number of BS antennas & $N=8\times 8$  \\
			\rowcolor{lightblue}
			Carrier frequency & 2.4 GHz\\
			Height of the BS & 10 m\\
			\rowcolor{lightblue}
			Spatial discretization unit & $\Delta_x=\Delta_y=1$\\
			Number of time slots & 64\\
			\rowcolor{lightblue}
			UT velocity & $0.6$ m/s \\
			Height of the UT & 1.5 m\\
			\rowcolor{lightblue}
			Batch size &64 \\
			Number of Laplacian pyramid levels & 3\\
			\rowcolor{lightblue}
			Number of WT levels & 2\\
			Number of Res-DSWT blocks in the low-frequency branch& $N_1=5$\\
			\rowcolor{lightblue}
			Number of Res-DSWT blocks in the high-frequency branch& $N_2=3$\\
			Initial learning rate& $\mathrm{lr}=2 \times 10^{-5}$\\
			\rowcolor{lightblue}
			Warm-up strategy (First 5,000 Iterations)& $2 \times 10^{-5} \to  1 \times 10^{-3}$\\
			Learning rate decay schedule&  50,000 iterations\\
			\rowcolor{lightblue}
			Proportion of zeros in ${\boldsymbol{\mathcal{D}}}$ & $p_{m}=0.2$\\
			Area of ${\mathcal{R}}_{\rm{inaccessible}}$ & $4\: \mathrm{m}\times4\: \mathrm{m}$\\
			\rowcolor{lightblue}
			Mask interval & $s=2$ m\\
			\bottomrule
		\end{tabular}
	}
\end{table}

\textit{iii) Training Strategy and Model Configuration:} At the hardware level, the proposed LPWTNet is trained on one Nvidia RTX-4090 GPU (24 GB memory) and tested on a single Nvidia RTX-4090 GPU with the same specifications. At the algorithmic level, we employ the Adam optimizer with an initial learning rate of $1 \times 10^{-3}$ to update the model parameters over 150,000 iterations, using a batch size of 64. To stabilize model training and reduce data-loading overhead, the dataset is enlarged tenfold by cyclically repeating the original data. During the first 5,000 iterations, a warm-up strategy is applied to ensure stable convergence at the early stage of training, where the learning rate is linearly increased from $2 \times 10^{-5}$ to $1 \times 10^{-3}$. Meanwhile, the learning rate follows a step-wise decay schedule, halving every 50,000 iterations. To ensure the generalization capability of LPWTNet, no checkpoint selection is performed, and only the most recent checkpoint is utilized. More detailed hyperparameter configurations of the proposed LPWTNet are provided in \tabref{ta:sys}.

\textit{iv) Evaluation Criteria:} For a fair comparison, we evaluate the accuracy of the proposed approach and baselines in constructing sCF using widely adopted metrics, namely normalized mean squared error (NMSE) and mean squared error (MSE). In addition, to assess the complexity of AI-based methods, we employ two commonly used measures: the number of model parameters to analyze storage complexity, and FLOPs to evaluate computational complexity.

\subsection{Experiment Results}
\textit{i) Ablation Study:} In this subsection, we verify the benefit of the enlarged receptive field introduced by the proposed DSWTConv for constructing the target sCF. Specifically, we investigate three variants of the proposed Res-DSWT module, including the full version, a variant in which the DSWTConv is replaced with a standard convolution, and another variant where the DSWTConv is replaced with a standard convolution further augmented by a self-attention mechanism \cite{vaswani2017attention}. Correspondingly, we evaluate three variants of the overall framework: the proposed full LPWTNet, LPWTNet with DSWTConv replaced by standard convolution (LPWTNet w/o DSWTConv), and LPWTNet with DSWTConv replaced by standard convolution and augmented with self-attention (LPWTNet w/o DSWTConv + SA). Meanwhile, this ablation study is conducted on the missing-region reconstruction task. 

\begin{figure}[!t]
	\centering
	\includegraphics[width=0.41\textwidth]{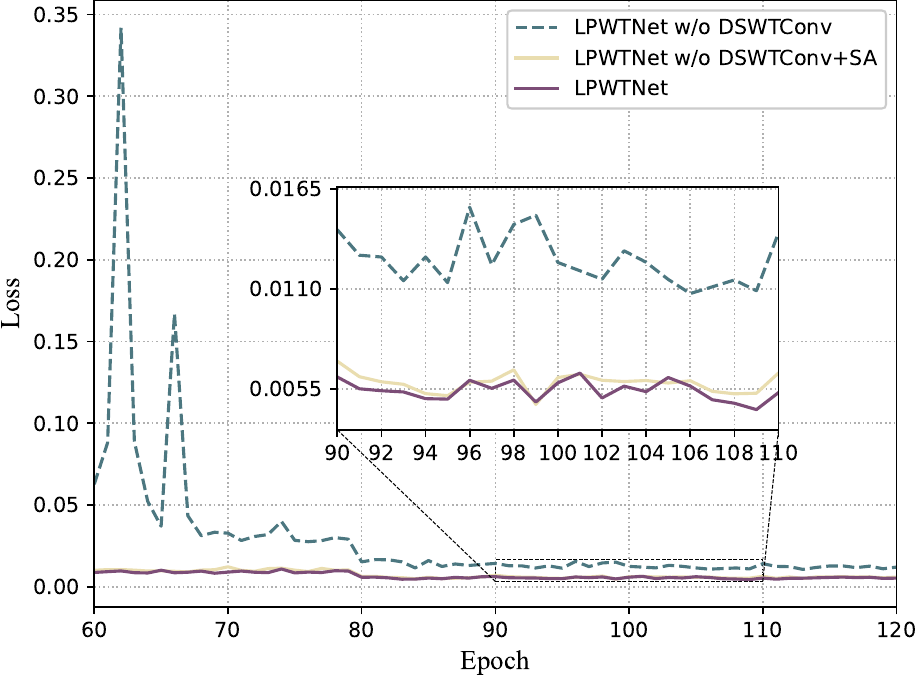}
	\captionsetup{font=footnotesize}
	\caption{Convergence analysis of different LPWTNet variants.}
	\label{fig:conver}
\end{figure}

As shown in \tabref{ta:abla}, we conduct a comprehensive comparison of the proposed DSWTConv module against standard convolution and the advanced self-attention mechanism in terms of both reconstruction accuracy and model complexity. The results show that LPWTNet with standard convolution exhibits the poorest performance due to the inherent limitations imposed by its restricted receptive field. Incorporating self-attention into LPWTNet yields performance gains; however, this comes at the cost of the highest FLOPs and parameter counts among all variants. In contrast, the proposed DSWTConv module not only makes a more substantial contribution to accurate sCF reconstruction but also achieves a remarkable reduction in both computational complexity (FLOPs in Giga) and storage complexity (parameters in Millions), with nearly a $50\%$ decrease compared to the other two modules. \figref{fig:conver} presents the convergence analysis of different LPWTNet variants in the later stages of training. The loss curve of LPWTNet w/o DSWTConv exhibits more pronounced fluctuations and remains at relatively higher values, indicating unstable and suboptimal convergence. In contrast, the full LPWTNet achieves faster and more stable convergence with consistently lower loss values. This observation further confirms that the enlarged receptive field provided by DSWTConv allows LPWTNet to capture richer contextual dependencies and higher-level semantic information, thereby guiding the optimization process toward better minima and accelerating overall convergence.

\begin{table*}[!t]
	\captionsetup{font=footnotesize}
	\caption{Ablation study of LPWTNet in the missing-region estimation task}\label{ta:abla}
	\centering
	\setlength{\tabcolsep}{2.1mm}
	\ra{2.1}
	\scriptsize
	\scalebox{0.9}{\begin{tabular}{LccccccR}
			\toprule
			Method & Depthwise Separable WTConv (DSWTConv)& Conv & Self-Attention (SA)& NMSE ($\times 10^{-3}$) & MSE ($\times 10^{-6}$) &FLOPs (G) & Parameters (M)   \\
			\midrule
			\rowcolor{lightblue}
			LPWTNet  & \checkmark& $\times$&$\times$ &\textbf{2.919} $\downarrow$   & \textbf{1.840} $\downarrow$   &\textbf{0.689} $\downarrow$&\textbf{13.508} $\downarrow$\\
			LPWTNet w/o DSWTConv  & $\times$& \checkmark& $\times$ &5.256 & 3.294  &1.132&22.077  \\
			\rowcolor{lightblue}
			LPWTNet w/o DSWTConv + SA  & $\times$ & \checkmark &\checkmark &3.861 & 2.435  &1.354&26.542 \\		
			\bottomrule
		\end{tabular}
	}
\end{table*}

\textit{ii) Performance Comparison with SOTA Baselines:} To ensure a fair comparison, we adopt the same training strategy as LPWTNet for all baselines, performing weight learning across the three different sCF construction tasks. To comprehensively evaluate the effectiveness and generalization capability of the proposed LPWTNet under different practical sCF measurement scenarios, we select several SOTA neural networks as benchmarks, including SRGAN-MSE \cite{ledig2017photo}, SRGAN \cite{ledig2017photo}, CVAE \cite{kingma2013auto}, and DRRN \cite{tai2017image}.

\tabref{Performance comparison-miss} presents the performance of the proposed LPWTNet compared with the above mentioned baselines under the missing-region estimation task. In terms of reconstruction accuracy, CVAE \cite{kingma2013auto} and DRRN \cite{tai2017image} outperform the GAN-based methods. Meanwhile, the proposed LPWTNet exhibits more competitive sCF reconstruction capability than both CVAE and DRRN, achieving lower NMSE and MSE values of $2.919\times10^{-3}$ and $1.840\times10^{-6}$, respectively. It is well known that the number of model parameters determines the storage complexity, where a larger parameter scale generally indicates a stronger representation capability. In contrast, FLOPs reflect the computational complexity and directly influence the inference speed; lower FLOPs typically correspond to faster computation and reduced hardware requirements. As shown in \tabref{Performance comparison-miss}, although the proposed approach contains a relatively large number of parameters, which inevitably increases the storage complexity, this design also equips the model with enhanced expressive power that contributes to high-precision sCF reconstruction. More importantly, the proposed LPWTNet achieves substantially lower computational complexity compared with the baseline methods. It requires only 0.689 G FLOPs, corresponding to approximately $2/5$ and $1/10$ of the computational cost of CVAE \cite{kingma2013auto} and DRRN \cite{tai2017image}, respectively. We attribute the ability of LPWTNet to simultaneously preserve strong representation capacity and maintain high computational efficiency to several key architectural designs. First, the LP and WTConv modules establish a multi-resolution and multi-frequency-aware computing paradigm, which enables more effective allocation of computational resources across different frequency components. Furthermore, in the overall network design, only the low-frequency sCF is forwarded into a deeper subnetwork branch, while the high-frequency sCFs are refined through lightweight mask-guided correction and a shallow subnetwork branch. These underlying design choices work synergistically to ensure that LPWTNet delivers high reconstruction accuracy while maintaining excellent computational efficiency. Moreover, the parameter count of the proposed model mainly arises from stacking multiple Res-DSWT blocks. By further assessing the relative importance of different Res-DSWT blocks in both the low- and high-frequency branches, and leveraging advanced importance-aware pruning techniques \cite{xu2023efficient,11220249,kim2024shortened} to enable structured model compression, the number of model parameters can be further reduced.

\begin{table}[!t]
	\captionsetup{font=footnotesize}
	\caption{Comparison of sCF construction performance between the proposed approach and baselines on the missing-region estimation task}\label{Performance comparison-miss}
	\centering
	\setlength{\tabcolsep}{0.9mm}
	\ra{2.1}
	\scriptsize
	\scalebox{0.9}{\begin{tabular}{LccccR}
			\toprule
			Method & NMSE ($\times 10^{-3}$) & MSE ($\times 10^{-6}$) & FLOPs (G) & Parameters (M)  \\
			\midrule
			\rowcolor{lightblue}
			SRGAN-MSE \cite{ledig2017photo} & 82.009 &  40.120 &1.155&6.248\\
			SRGAN \cite{ledig2017photo}   & 146.150  & 81.261 &2.250& 20.923\\
			\rowcolor{lightblue}
			CVAE \cite{kingma2013auto}   & 12.836  & 8.012 &1.739 & \textbf{1.697} $\downarrow$\\
			DRRN \cite{tai2017image}       & 14.571 & 10.405& 6.191& 6.046\\
			\rowcolor{lightblue}
			LPWTNet (Our)  & \textbf{2.919} $\downarrow$  & \textbf{1.840} $\downarrow$ &\textbf{0.689} $\downarrow$&13.508\\

			\bottomrule
		\end{tabular}
	}
\end{table}

As shown in \tabref{Performance comparison-unifor}, we compare the proposed LPWTNet with the baselines on the uniform sparse recovery task. It can be observed that SRGAN-MSE \cite{ledig2017photo} exhibits relatively poor reconstruction performance, while SRGAN \cite{ledig2017photo} and CVAE \cite{kingma2013auto} achieve comparable results. In contrast, the proposed LPWTNet consistently delivers more favorable performance across both accuracy- and computation-related metrics. \tabref{Performance comparison-nonunifor} further presents the performance comparison on the non-uniform sparse recovery task. A similar trend is observed: the proposed LPWTNet demonstrates substantially enhanced sCF reconstruction capability, whereas the GAN-based baselines tend to yield inferior results. When considering all three reconstruction tasks collectively, the proposed LPWTNet consistently delivers competitive and robust performance across diverse sCF degradation scenarios. In addition, CVAE \cite{kingma2013auto} and DRRN \cite{tai2017image} generally exhibit more stable reconstruction behavior than SRGAN-MSE \cite{ledig2017photo} and SRGAN \cite{ledig2017photo}.

\begin{table}[!t]
	\captionsetup{font=footnotesize}
	\caption{Comparison of sCF construction performance between the proposed approach and baselines on the uniform sparse recovery task}\label{Performance comparison-unifor}
	\centering
	\setlength{\tabcolsep}{0.9mm}
	\ra{2.1}
	\scriptsize
	\scalebox{0.9}{\begin{tabular}{LccccR}
			\toprule
			Method & NMSE ($\times 10^{-3}$) & MSE ($\times 10^{-6}$) & FLOPs (G) & Parameters (M)  \\
			\midrule
			\rowcolor{lightblue}
			SRGAN-MSE \cite{ledig2017photo} & 125.067 & 61.517 &1.155&6.248\\
			SRGAN \cite{ledig2017photo}   & 52.314  & 29.337 &2.250& 20.923\\
			\rowcolor{lightblue}
			CVAE \cite{kingma2013auto}   & 54.801  & 33.053 &1.739 & \textbf{1.697} $\downarrow$\\
			DRRN \cite{tai2017image}       & 32.691 & 17.999& 6.191& 6.046\\
			\rowcolor{lightblue}
			LPWTNet (Our) & \textbf{22.946} $\downarrow$  & \textbf{12.406} $\downarrow$ &\textbf{0.689} $\downarrow$&13.508\\

			\bottomrule
		\end{tabular}
	}
\end{table}
\begin{table}[!b]
	\captionsetup{font=footnotesize}
	\caption{Comparison of sCF construction performance between the proposed approach and baselines on the non-uniform sparse recovery task}\label{Performance comparison-nonunifor}
	\centering
	\setlength{\tabcolsep}{0.9mm}
	\ra{2.1}
	\scriptsize
	\scalebox{0.9}{\begin{tabular}{LccccR}
			\toprule
			Method & NMSE ($\times 10^{-3}$) & MSE ($\times 10^{-6}$) & FLOPs (G) & Parameters (M)  \\
			\midrule
			\rowcolor{lightblue}
			SRGAN-MSE \cite{ledig2017photo} & 123.527 & 72.647 &1.155&6.248\\
			SRGAN \cite{ledig2017photo}   & 133.320  & 78.609 &2.250& 20.923\\
			\rowcolor{lightblue}
			CVAE \cite{kingma2013auto}   & 7.314  & 4.496 &1.739 & \textbf{1.697} $\downarrow$\\
			DRRN \cite{tai2017image}       & 87.832 & 58.353 & 6.191& 6.046\\
			\rowcolor{lightblue}
			LPWTNet (Our)  & \textbf{3.220} $\downarrow$  & \textbf{2.100} $\downarrow$ &\textbf{0.689} $\downarrow$&13.508\\

			\bottomrule
		\end{tabular}
	}
\end{table}

We attribute the relatively competitive performance of DRRN to two key factors. First, its objective function is explicitly formulated to minimize the reconstruction error, directly enforcing fidelity between the predicted and ground-truth sCFs. Second, the residual learning mechanism together with a deep recursive architecture substantially enhances its representation capacity, enabling more accurate detail recovery \cite{tai2017image}. In contrast, the objective of VAE-based approaches is to optimize the evidence lower bound (ELBO). Since this loss corresponds to a well-defined variational inference objective under the maximum likelihood estimation framework, VAE training typically exhibits superior stability \cite{kingma2013auto}. Moreover, CVAE leverages an additional MSE loss computed in the original sCF domain, together with a likelihood maximization objective in the latent space, thereby further improving the reconstruction fidelity. GAN-based methods, on the other hand, rely on adversarial training to jointly optimize a generator and a discriminator with the goal of reaching a Nash equilibrium. However, the inherently dynamic nature of this two-player optimization often leads to unstable convergence, which may manifest as oscillatory loss behavior or even mode collapse \cite{cao2024survey}. Moreover, the primary design objective of GANs is not to minimize reconstruction-oriented metrics such as MSE but rather to enhance perceptual realism. As a result, GAN-based approaches may generate visually plausible yet structurally inaccurate details, making them more suitable for general generative tasks than for precise quantitative estimation.
\newcolumntype{L}{>{\hspace*{-\tabcolsep}}l}
\newcolumntype{R}{c<{\hspace*{-\tabcolsep}}}
\definecolor{lightblue}{rgb}{0.93,0.95,1.0}
\begin{table*}[!t]
	\captionsetup{font=footnotesize}
	\caption{Zero-shot performance comparison of the proposed approach and baselines on sCF reconstruction for the non-uniform sparse recovery and uniform sparse recovery tasks}\label{zero-shot}
	\centering
	\setlength{\tabcolsep}{2.6mm}
	\ra{2.1}
	\scriptsize
	\begin{tabular}{LcccccR}
		\toprule
		\multirow{2}{*}{Method} 
		&\multicolumn{2}{c}{\begin{tabular}[c]{@{}c@{}}Non-Uniform Sparse\\ Recovery Task (Zero-Shot)\end{tabular}} 
		&\multicolumn{2}{c}{\begin{tabular}[c]{@{}c@{}}Uniform Sparse\\Recovery Task (Zero-Shot)\end{tabular}} 
		&\multirow{2}{*}{FLOPs (G)} 
		&\multirow{2}{*}{Parameters (M)} \\
		&\multicolumn{1}{c}{NMSE ($\times 10^{-3}$)} 
		&\multicolumn{1}{c}{MSE ($\times 10^{-6}$)} 
		&\multicolumn{1}{c}{NMSE ($\times 10^{-3}$)} 
		&\multicolumn{1}{c}{MSE ($\times 10^{-6}$)}\\
		\midrule
		\rowcolor{lightblue}
		\multicolumn{1}{L}{SRGAN-MSE \cite{ledig2017photo}}
		& \multicolumn{1}{c}{221.162} & \multicolumn{1}{c}{128.981} & \multicolumn{1}{c}{187.537} & \multicolumn{1}{c}{114.284}& \multicolumn{1}{c}{1.155}&\multicolumn{1}{R}{6.248} \\
		
		\multicolumn{1}{L}{SRGAN \cite{ledig2017photo}}
		& \multicolumn{1}{c}{245.724} & \multicolumn{1}{c}{138.281} & \multicolumn{1}{c}{287.467} & \multicolumn{1}{c}{175.181} & \multicolumn{1}{c}{2.250} & \multicolumn{1}{R}{20.923} \\
		
		\rowcolor{lightblue}
		\multicolumn{1}{L}{CVAE \cite{kingma2013auto}}
		& \multicolumn{1}{c}{41.562} & \multicolumn{1}{c}{28.238} & \multicolumn{1}{c}{134.570} & \multicolumn{1}{c}{91.572} & \multicolumn{1}{c}{1.739} & \multicolumn{1}{R}{\textbf{1.697} $\downarrow$} \\
		
		\multicolumn{1}{L}{DRRN \cite{tai2017image}}
		& \multicolumn{1}{c}{98.901} & \multicolumn{1}{c}{65.952} & \multicolumn{1}{c}{249.745} & \multicolumn{1}{c}{166.024} & \multicolumn{1}{c}{6.191} & \multicolumn{1}{R}{6.046} \\
		
		\rowcolor{lightblue}
		\multicolumn{1}{L}{CGDM \cite{11220249}}
		&\multicolumn{1}{c}{12.619} 
		&\multicolumn{1}{c}{8.416} 
		&\multicolumn{1}{c}{86.812} 
		&\multicolumn{1}{c}{57.369} 
		&\multicolumn{1}{c}{54.959} 
		&\multicolumn{1}{R}{3539.120} \\
		
		\multicolumn{1}{L}{LPWTNet (Our)}
		& \multicolumn{1}{c}{\textbf{9.015} $\downarrow$} & \multicolumn{1}{c}{\textbf{5.8261} $\downarrow$} & \multicolumn{1}{c}{\textbf{62.0128} $\downarrow$} & \multicolumn{1}{c}{\textbf{40.218} $\downarrow$} & \multicolumn{1}{c}{\textbf{0.689} $\downarrow$} & \multicolumn{1}{R}{13.508} \\
		
		\bottomrule
	\end{tabular}
\end{table*}
\newcolumntype{L}{>{\hspace*{-\tabcolsep}}l}
\newcolumntype{R}{c<{\hspace*{-\tabcolsep}}}
\definecolor{lightblue}{rgb}{0.93,0.95,1.0}
\begin{table*}[!b]
	\captionsetup{font=footnotesize}
	\caption{Performance comparison for sCF reconstruction across different tasks in the ``3GPP 38.901 Indoor LOS'' wireless communication scenario}\label{los_indoor_zero-shot}
	\centering
	\setlength{\tabcolsep}{2.6mm} 
	\ra{2.1} 
	\scriptsize
	\begin{tabular}{LcccccR}
		\toprule
		\multirow{2}{*}{Method} 
		
		& \multicolumn{2}{c}{\begin{tabular}[c]{@{}c@{}}Missing-Region\\Estimation Task\end{tabular}} 
		
		& \multicolumn{2}{c}{\begin{tabular}[c]{@{}c@{}}Non-Uniform Sparse\\Recovery Task (Zero-Shot)\end{tabular}} 
		
		& \multicolumn{2}{c}{\begin{tabular}[c]{@{}c@{}}Uniform Sparse\\Recovery Task (Zero-Shot)\end{tabular}} \\
		
		& NMSE ($\times 10^{-3}$) 
		& MSE ($\times 10^{-6}$) 
		
		& NMSE ($\times 10^{-3}$) 
		& MSE ($\times 10^{-6}$) 
		
		& NMSE ($\times 10^{-3}$) 
		& MSE ($\times 10^{-6}$) \\
		
		\midrule
		
		\rowcolor{lightblue}
		SRGAN-MSE \cite{ledig2017photo}
		& 288.088 & 135.531
		& 267.508 & 132.174
		& 217.234  & 130.608\\
		
		SRGAN \cite{ledig2017photo}
		& 161.121 & 83.024
		& 185.477 & 97.220
		& 248.744 & 150.644 \\
		
		\rowcolor{lightblue}
		CVAE \cite{kingma2013auto}
		& 6.157 & 3.052
		& 35.473 & 21.016 
		& 120.597 & 83.372 \\
		
		DRRN \cite{tai2017image}
		& 6.989 & 3.459
		& 92.522 & 59.413
		& 198.515 & 123.360 \\
		
		\rowcolor{lightblue}
		CGDM \cite{11220249}
		& 6.002 & 2.971
		& 12.016 & 7.716
		& 28.968 & 13.510  \\
		
		LPWTNet (Our)
		& \textbf{2.013} $\downarrow$ & \textbf{1.021} $\downarrow$
		& \textbf{5.502} $\downarrow$ & \textbf{2.887} $\downarrow$
		& \textbf{20.693} $\downarrow$ & \textbf{9.779} $\downarrow$ \\
		
		\bottomrule
	\end{tabular}
\end{table*}

\textit{iii) Evaluation of Knowledge Transfer and Generalization Ability:} Transferring a trained model to unseen tasks is regarded as zero-shot testing. Specifically, the neural network is trained and updated exclusively on the dataset for the missing-region estimation task, without any exposure to the datasets of the uniform sparse recovery and non-uniform sparse recovery tasks. The trained model is then directly applied to perform sCF reconstruction for different tasks that have not been encountered during training. This setting provides an effective way to evaluate the model's knowledge transferability and generalization capability. Furthermore, to enable a more comprehensive evaluation of reconstruction accuracy and generalization performance in sCF reconstruction tasks, a representative deep generative model, namely the conditional generative diffusion model (CGDM) \cite{11220249}, is additionally included as one of the baselines. \tabref{zero-shot} summarizes the zero-shot performance of the proposed LPWTNet and the baseline methods on both uniform sparse recovery and non-uniform sparse recovery tasks. It can be observed that, across the two unseen sCF reconstruction tasks, SRGAN and SRGAN-MSE exhibit slightly inferior performance, while CGDM and CVAE achieve relatively better results. We attribute this to CGDM's robust ability to learn implicit priors and its capacity to model complicated data distributions. Notably, the proposed approach achieves more competitive NMSE and MSE performance across multiple unseen sCF reconstruction tasks, further demonstrating its superior knowledge transferability and generalization capability.

To further assess the generalization capability of the proposed approach across different wireless communication scenarios, additional experiments are conducted in a typical 5G NR indoor office scenario, namely the ``3GPP 38.901 Indoor LOS'' scenario, along with zero-shot evaluations on unseen task datasets. For this indoor scenario, the preprocessing procedure of the sCF data follows the same settings as in the previously considered scenarios. As shown in \tabref{los_indoor_zero-shot}, the proposed approach and baseline models are trained on the dataset corresponding to the missing-region estimation task, and the resulting models are directly applied to the unseen datasets of the non-uniform sparse recovery and uniform sparse recovery tasks for zero-shot evaluation. A consistent performance trend is observed across different sCF reconstruction tasks: GAN-based methods exhibit relatively inferior performance, whereas CGDM, CVAE, and DRRN achieve comparatively better results. In particular, CGDM demonstrates strong generalization capability in zero-shot testing. Importantly, the proposed LPWTNet consistently achieves more competitive sCF reconstruction performance while maintaining a more favorable trade-off in computational complexity. These results further underscore the effectiveness and robustness of the proposed approach across diverse wireless communication scenarios and tasks.

\section{Conclusion}\label{sec:conclusion}
In this paper, we investigated the construction of sCFs for massive MIMO systems and introduced a unified tensor-based learning architecture to address a variety of practical scenarios. By modeling the sCF as a tensor and exploiting the correlation between the CSCM and the CPAS, we effectively reduced the dimension of sCF representation. Moreover, we uniformly formulated multiple sCF construction tasks under different measurement constraints as tensor restoration problems and developed LPWTNet to solve them efficiently. Specifically, a parameter-free LP was introduced to achieve multi-scale frequency decomposition, while a shared mask learning strategy enabled level-wise refinement. Additionally, a Res-DSWT block was designed to enlarge the receptive field and could be flexibly integrated into existing neural network architectures. Experimental results demonstrated that the proposed framework achieved substantial performance improvements over the baseline methods across various wireless communication scenarios and representative tasks.

\bibliographystyle{IEEEtran}
\bibliography{EE_AI}

\end{document}